\documentclass[lettersize,journal]{IEEEtran}
\usepackage{amsmath,amsfonts}
\usepackage{algorithmic}
\usepackage{algorithm}
\usepackage{array}

\usepackage{textcomp}
\usepackage{stfloats}
\usepackage{url}
\usepackage{verbatim}
\usepackage{graphicx}
\usepackage{cite}
\usepackage{bbding}

\usepackage{graphicx}
\usepackage{subfigure}
\usepackage[backref]{hyperref}
\usepackage{multirow}
\usepackage{multicol}
\usepackage{float}
\usepackage{caption}
\usepackage{wrapfig}
\usepackage{enumerate}
\usepackage{fancyhdr}

\usepackage{amssymb}
\usepackage{color}
\usepackage{listings}

\begin{document}

\title{Universum-inspired Supervised Contrastive Learning}

\author{Aiyang~Han,~Chuanxing~Geng,~Songcan~Chen\IEEEauthorrefmark{2}% <-this % stops a space
\thanks{A. Han, C. Geng and S. Chen are with Nanjing University of Aeronautics and Astronautics.}%
\thanks{\IEEEauthorrefmark{2}Corresponding author: s.chen@nuaa.edu.cn}% <-this % stops an unwanted space 
\thanks{This paper has supplementary downloadable material available at \href{http://ieeexplore.ieee.org}{http://ieeexplore.ieee.org.}, provided by the author. The material includes a supplemental essay on the details of the experiments as well as the algorithm. Contact s.chen@nuaa.edu.cn for further questions about this work.}
}

\markboth{Journal of \LaTeX\ Class Files,~Vol.~14, No.~8, August~2021}%
{Han, Geng and Chen \MakeLowercase{\textit{et al.}}: Universum-inspired Supervised Contrastive Learning}


\maketitle
\thispagestyle{fancy}
\lhead{}
\lfoot{}
\cfoot{\tiny{This article has been accepted for publication in IEEE Transactions on Image Processing. This is the author's version which has not been fully edited and
content may change prior to final publication. Citation information: DOI 10.1109/TIP.2023.3290514. \copyright~2023 IEEE. Personal use is permitted, but republication/redistribution requires IEEE permission.
See https://www.ieee.org/publications/rights/index.html for more information.}}
\rfoot{}

\begin{abstract}
As an effective data augmentation method, Mixup synthesizes an extra amount of samples through linear interpolations. Despite its theoretical dependency on data properties, Mixup reportedly performs well as a regularizer and calibrator contributing reliable robustness and generalization to deep model training. In this paper, inspired by Universum Learning which uses out-of-class samples to assist the target tasks, we investigate Mixup from a largely under-explored perspective - the potential to generate in-domain samples that belong to none of the target classes, that is, \emph{universum}. We find that in the framework of supervised contrastive learning, Mixup-induced universum can serve as surprisingly high-quality hard negatives, greatly relieving the need for large batch sizes in contrastive learning. With these findings, we propose \underline{\textbf{Uni}}versum-inspired supervised \underline{\textbf{Con}}trastive learning (UniCon), which incorporates Mixup strategy to generate \emph{Mixup-induced universum} as universum negatives and pushes them apart from anchor samples of the target classes. We extend our method to the unsupervised setting, proposing \underline{\textbf{Un}}supervised \underline{\textbf{Uni}}versum-inspired contrastive model (Un-Uni). Our approach not only improves Mixup with hard labels, but also innovates a novel measure to generate universum data. With a linear classifier on the learned representations, UniCon shows state-of-the-art performance on various datasets. Specially, UniCon achieves 81.7\% top-1 accuracy on CIFAR-100, surpassing the state of art by a significant margin of 5.2\% with a much smaller batch size, typically, 256 in UniCon vs. 1024 in SupCon\cite{supcon} using ResNet-50. Un-Uni also outperforms SOTA methods on CIFAR-100. The code of this paper is released on \href{https://github.com/hannaiiyanggit/UniCon}{https://github.com/hannaiiyanggit/UniCon}.
\end{abstract}


\section{Introduction}\label{sec:introduction}

\IEEEPARstart{A}{s} a strong augmentation technique in supervised learning, Mixup has empirically and theoretically been proved to boost the performance of neural networks with its regularization power \cite{mixup,mixup1,mixup2}. Besides its reliable performance, Mixup is also reported to strengthen deep models with better calibration \cite{mixup3}, robustness \cite{mixup4,mixup5} and generalization \cite{mixup4}, thus being widely used in adversarial training \cite{mixup2}, domain adaptation \cite{mixup6}, imbalance problems \cite{mixup7} and so on. However, as Mixup-style training depends heavily on data properties\cite{mixup8}, on certain cases, chances are that traditional Mixup labels cannot correctly describe the augmented data. These labels, when taken as the ground truth, may provide \emph{unreliable supervision} for learners.

\begin{figure*}[htbp]
\includegraphics[width=\textwidth]{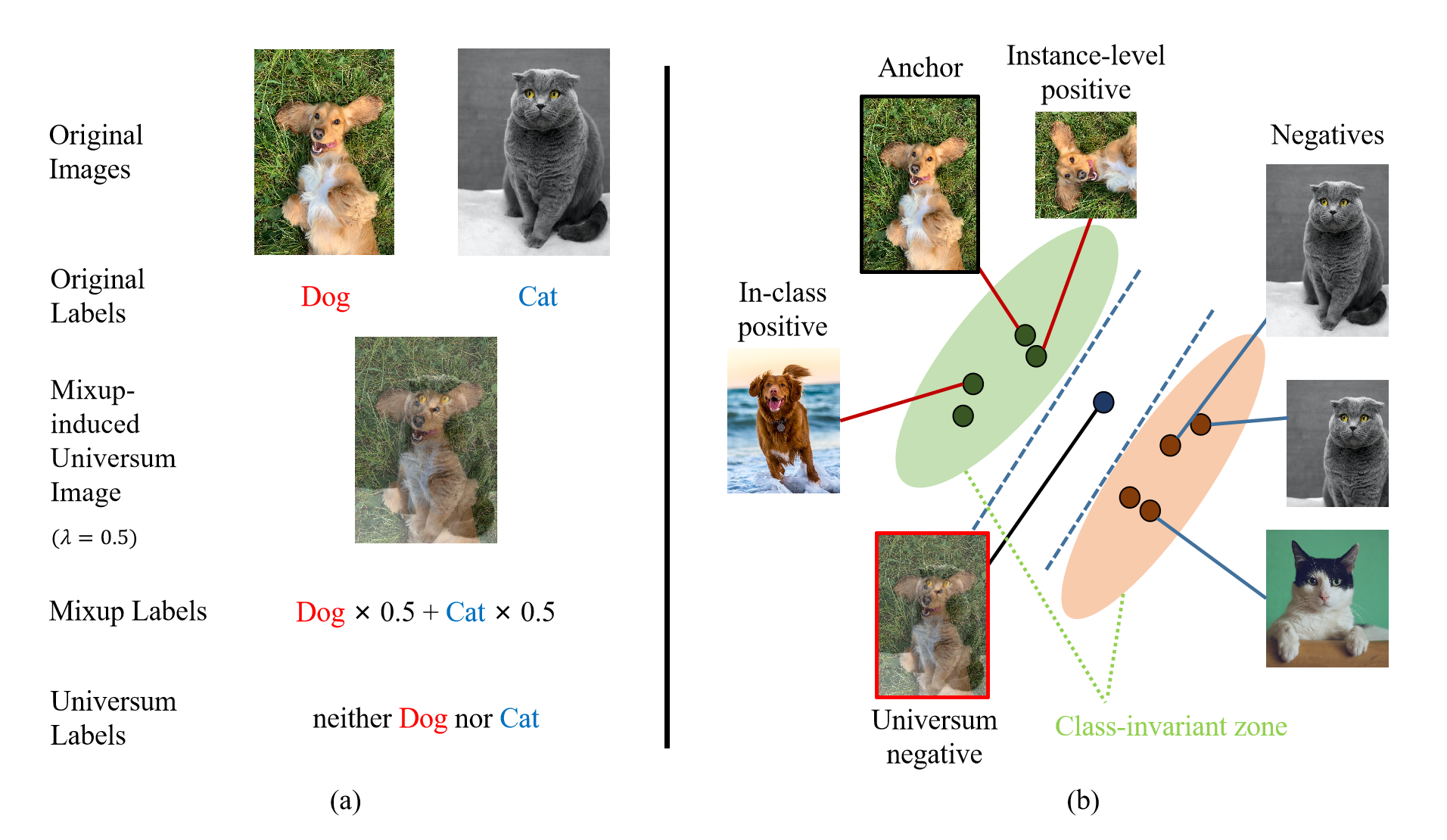}
\centering
\caption{The intuition behind our model. \textbf{(a)}: When processing Mixup labels, traditional method uses the mixture of original labels, but universum-style method regards Mixup data points as belonging to neither of the original classes, thus assigning the new points to a generalized negative class which is compulsorily limited to some desired region. \textbf{(b)}: In the framework of supervised contrastive learning, universum-style Mixup images can serve as negative samples for all anchor samples of the target classes. By pushing these Mixup-induced universum (universum negatives) apart from other data points, the model can better separate images from different classes. } \label{fig1}
\end{figure*}
To solve the problem of unreliable labelling, Universum learning allows us to see Mixup in a new light. Introduced by \cite{universum,universum2}, universum is referred to as in-domain samples that belong to none of the target classes in classification. In the scenario of universum learning, usually a new dataset of universum is introduced to assist classification of the target dataset (e.g. hand-written letters are introduced to help classify hand-written digits) \cite{ulda,upractical,usvm}. Although universum data cannot be assigned to the classes in question, they still can be constructed into a regularization term so as to improve the model performance with their domain knowledge and negativity \cite{universum2}. From the perspective of universum learning, here comes a natural question: instead of using the linear interpolations of original labels, why don't we assign Mixup samples to a generalized negative class? Just as humans may perceive, if an animal is half dog and half cat, it is actually of neither species. The conventional methods of Mixup treat Mixup data from various Mixup (or combined) coefficients equally, while, in reality, different coefficients could make data of different characteristics. This paper intends to argue that there exist some special values of coefficient $\lambda$, especially $\lambda=0.5$, that generate a bundle of Mixup data which can be hardly related to any of the semantics of the original images so as to possess some characteristics of universum data. Therefore, these Mixup data are denoted as \emph{Mixup-induced universum}.

As is shown in Fig.~\ref{fig1}(a), Mixup-induced universum (the Mixup image) is regarded as neither dog nor cat, but rather an \emph{universum} data point. With this approach, models can be free from the concern of unreliable ground truth labels in Mixup. What's more, the combination of universum learning and Mixup also introduces \emph{a new way to acquire universum data}, which extends universum learning to fully-supervised setting. Compared with foreign samples such as hand-written letters in the classification of hand-written digits, universum data produced by Mixup are semantically closer to target data, which may provide better regularization effects in training.

The na\"ive way of assigning Mixup-induced universum to a newly defined category may result into imbalanced data when Mixup data far outweigh original data in amount. Therefore, to benefit more from large amounts of universum data, a contrastive framework is adopted in this paper. Recently, contrastive learning has greatly boosted deep learning via pulling together positive sample pairs and separating negative pairs in the embedding space\cite{simclr,moco,swav,supcon,barlow}. Early contrastive models only take augmentations of the same image as positive pairs, while treating all other sample pairs as negative pairs\cite{simclr,moco}. Specially, SupCon model extended contrastive learning to the fully-supervised setting by including samples from the same class into positives for each anchor sample \cite{supcon}.

Although contrastive learning and Mixup both improve the performance of supervised learning, the combination of the two can be especially difficult due to their opposite ways of organizing data. While Mixup softly assigns augmented data to multiple classes \cite{cutmix}, contrastive learning requires hard labels to compute the contrastive loss. A few attempts have been made to conjoin contrastive learning and Mixup either by designing a Mixup version of InfoNCE loss \cite{mixco} or by using the na\"ive addition of the InfoNCE loss and the Mixup-style cross entropy loss\cite{covid}. A better exploration might be MoCHi \cite{hardmix}, which applies Mixup only to the hard negatives in the memory bank so as to acquire more and harder negatives. However, these methods pay more attention to softening the contrastive learning rather than innovating Mixup strategy, ignoring the innate potential of Mixup to produce negative samples.

In this paper, inspired by universum learning, we introduce a novel measure to combine contrastive learning and Mixup with the simple idea that \emph{Mixup samples could be hard negatives}. Unlike \cite{hardmix} that selects and mixes hard negatives, our method randomly mixes two images from different classes and assumes that these Mixup data are hard due to their visual ambiguity. Following the framework of supervised contrastive learning, we go a step further to include Mixup images into the contrastive loss by viewing them as \emph{Mixup-induced universum} - universum data which are negative to the global dataset - in contrast with traditional negatives that are negative for a limited group of anchor samples. As is shown in Fig. \ref{fig1}(b), we incorporates Mixup to generate Mixup-induced universum and pushes them apart from anchor samples of the target classes. For each anchor sample, a contrast sample is chosen from other classes to synthesize a universum data point, which helps establish clearer margins among different instances as well as different classes. Since traditional Mixup strategy that samples the Mixup parameter from Beta distribution \cite{mixup} may generate samples semantically close to a target class, we fix the Mixup parameter to a constant, thereby driving the synthesized universum data out of the regions of target classes in the data space. Although the idea is simple, there is no prior knowledge on how to contrast these universum negatives with anchor samples. We design two loss functions based on the intuitions mentioned above, and empirically show that an entirely universum-based loss achieves better performance on datasets. In such a universum-based framework, universum data are adopted both for contrast with negatives and derivation of class centers. Despite the coarse design of this loss function, it is especially effective with universum data used in all stages to help construct more robust and representative features. Our use of universum data spares us the efforts for hard negative mining, as Mixup samples naturally become hard negatives with their visual ambiguity.

Our work provides an effective method for fully-supervised learning. We validate the performance of UniCon on a range of datasets. On ResNet-50 \cite{resnet}, UniCon achieves 81.7\% top-1 accuracy on CIFAR-100 and 97.4\% on CIFAR-10 \cite{cifar}, which surpasses the state of art\cite{supcon} by 5.2\% and 1.4\% respectively. Our method can be applied to other contrastive learning methods in need of large amounts of negatives. This paper is based on our APWeb-WAIM paper\cite{ours} and extended in several aspects:

\begin{enumerate}[i)]
    \item We theoretically and empirically prove that our proposed loss function can benefit from hard universum negatives, while contributing to large margins among different classes.
    \item We test UniCon on CIFAR-100-C and TinyImageNet-C to prove its robustness. Our proposed method greatly strengthens model robustness in the face of various corruptions.
    \item We conduct a comprehensive experiment to explore the performance of SOTA models combining augmentations and two mixture methods (Mixup\cite{mixup} and CutMix\cite{cutmix}). It is shown that UniCon outperforms other models even when applied with the exactly same tricks.
    \item We newly propose the unsupervised version of our model, which, on the basis of UniCon, is achieved by simply using data points' indices in a batch for their pseudo-labels.
\end{enumerate}

Our main contributions are as follows:
\begin{itemize}
\item[$\bullet$] We investigate Mixup from the perspective of universum learning, thus unearthing Mixup's potential of generating samples that lie in the same domain as the target data yet belong to none of the target classes. We dig out Mixup as a novel measure to acquire universum data from a fully supervised dataset.
\item[$\bullet$] We introduce \underline{\textbf{Uni}}versum-inspired supervised \underline{\textbf{Con}}trastive learning (UniCon), which incorporates Mixup to generate Mixup-induced universum as negatives and pushes them apart from anchor samples of the target classes. Different from other contrastive models where the negativity of samples varies with anchors, such universum negatives in our model are negative to the global dataset. To our best knowledge, this is the first time that Mixup is used to produce a generalized negative class.
\item[$\bullet$] We find that in the framework of supervised contrastive learning, Mixup samples can work surprisingly good as hard negatives.
\item[$\bullet$] We show that our model can achieve outstanding performance on a range of datasets with a relatively small-scale neural network as well as a smaller batch size.
\item[$\bullet$] In the unsupervised setting, our proposed \underline{\textbf{Un}}supervised \underline{\textbf{Uni}}versum-inspired contrastive model (Un-Uni) also achieves state-of-the-art performance.
\end{itemize}

\section{Related Works}\label{sec:Related Works}
In this section, we will give a brief introduction of Mixup, universum learning and contrastive learning, as well as their relation to our method.
\subsection{Mixup}
Since Mixup was proposed by \cite{mixup}, it has been widely accepted as an effective and efficient measure for deep training \cite{mixup1,mixup2}. Despite Mixup's outstanding performance, recently the foundations of Mixup have also been scrutinized in theory. \cite{mixup1} theoretically proves that Mixup is a strong regularizer and equals to a standard empirical risk minimization estimator in the face of noises. \cite{mixup3} focuses on Mixup's effects of improving calibration and predictive uncertainty. \cite{mixup4} gives a theoretical explanation on how Mixup contributes to robustness and generalization of deep models. While Mixup is empirically and theoretically proved a reliable method, \cite{mixup8} demonstrates its data dependency by computing a closed form for the Mixup-optimal classification, and thereby providing a failure case of Mixup. This failure case indicates that Mixup could also be misleading as the synthesized data points are still softly connected with the original labels. Our method intends to disconnect the Mixup data from all known classes so that the additional domain knowledge could be learned without misleading information.
\subsection{Universum Learning}
Universum was introduced by Vapnik as ``an alternative capacity concept to the large margin approach", which indicates a group of samples that cannot be assigned to any target class in classification \cite{universum}. Universum learning is mostly explored as a new research scenario where a relevant dataset is introduced to assist the tasks on the target dataset. \cite{universum2} has theoretically proved that the use of universum data could benefit Support Vector Machines (SVM) with regularization effects. Various research has extended Universum Learning to metric learning \cite{umetric}, canonical correlation analysis \cite{ucca}, transductive learning \cite{transuni} and so on. By using unlabeled data as universum data, \cite{unipres} theoretically and empirically proves the efficiency of such universum prescription. Inspired by universum learning, our model, instead of importing a dataset, generates a group of universum samples from the target dataset to assist classification.
\subsection{Contrastive Learning}
Contrastive learning learns deep representations through contrasting positive sample pairs against negative ones. The definition of positive and negative pairs varies with different contrastive models. SimCLR \cite{simclr, simclrv2} and MoCo \cite{moco, mocov2} only admit augmentations of the same image as positive pairs, while cluster-based methods like SupCon \cite{supcon} and SwAV \cite{swav} also give in-class positives a pass. While classical contrastive models use the InfoNCE loss \cite{cpc}, more contrastive losses have flourished \cite{deepcluster, infomin, infomax}. For example, Barlow Twins \cite{barlow} aims to reduce data redundancy with a cross-correlation matrix, while BYOL \cite{byol} strengthens the consistency among views by predicting the second view from the first one. For further details on contrastive learning, we refer our readers to \cite{sslsurvey}.

Several attempts have been made to construct Mixup-style contrastive models \cite{mixco,covid,unmix,hardmix, enaet, neighborhoodcontrastive}. Mixco \cite{mixco} pulls Mixup data towards their original images in a Mixup way, while MoCHi \cite{hardmix} uses Mixup only on the hard negatives to capture the hardest negatives. Similarly, \cite{neighborhoodcontrastive} provides a semi-supervised version of MoCHi. In the unsupervised setting, Un-Mix \cite{unmix} only mixes the images, closing the distance among the Mixup image and an augmented version of Mixup data. Different from them, UniCon does not combine Mixup and contrastive learning in a na\"ive way. Instead, we delve into the nature of hard negatives, adopting Mixup as a way of hard negative generation. In this way, we not only train a more effective model, but also relieve the need for a large batch size in contrastive learning as is shown in the latter experimental results.

\section{Method}\label{method}
This section begins with a brief introduction of self-supervised and supervised contrastive losses, after which we present universum-style Mixup method. Then, with the Mixup-induced universum, small-scale and large-scale UniCon losses are proposed, while the latter is empirically proved to be a better one.

\begin{figure}[htbp]
\includegraphics[width=0.5\textwidth]{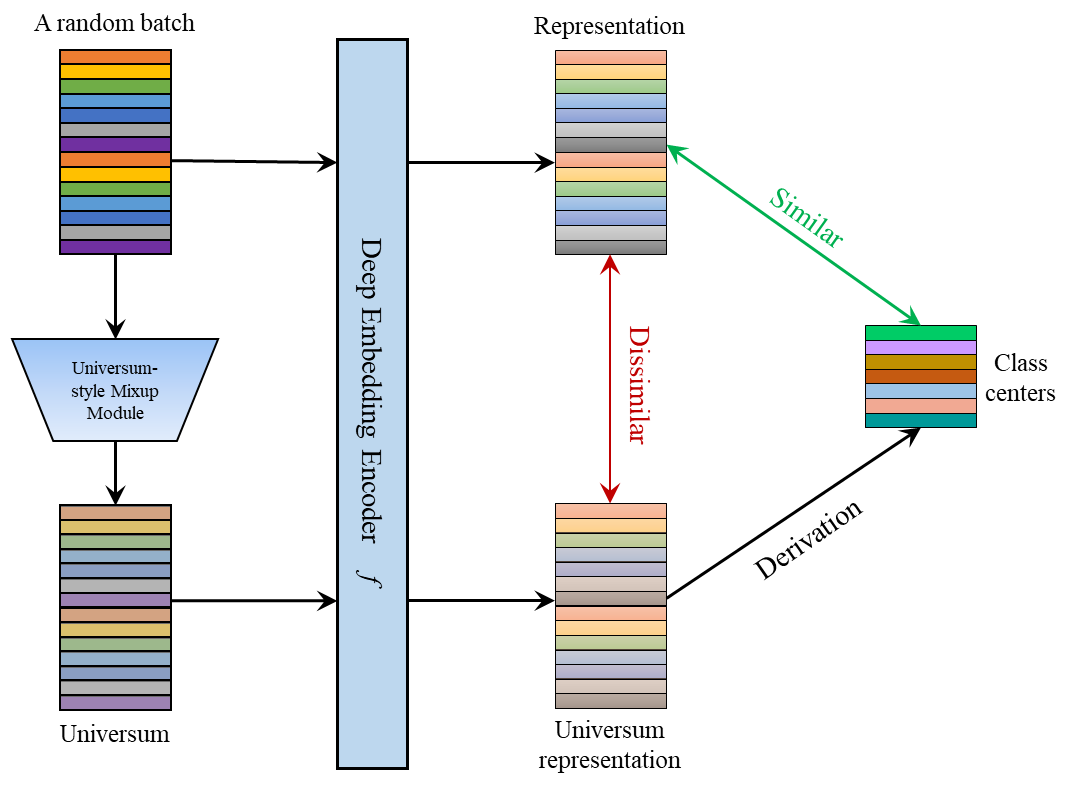}
\centering
\caption{An overview of UniCon. First a random batch is put through universum-style Mixup module to produce a batch of universum. Then both the original batch and universum data are encoded into deep representations, while universum representations are further utilized to generate class centers. After that, the model maximizes the similarity between anchors and their corresponding class centers while minimizing the similarity between anchors and all universum data points.} \label{fig2}
\end{figure}

Following the framework of \cite{supcon}, our approach is in nature a representation learning method. A deep encoder $f$ is adopted to learn the representations of target samples through minimizing a proposed loss. With $N$ being the batch size, each data point and its label are denoted by $x_{k}$ and $y_{k}$ $(k=1,2,..,N)$, while the corresponding augmented sample and its label is denoted by $\widetilde x_{k}$ and $\widetilde y_{k}$ $(k=1,2,..,2N)$. Note that $\widetilde x_{2k-1}$ and $\widetilde x_{2k}$ are two transformed augmentations derived from $x_{k}$, thus $\widetilde y_{2k-1} = \widetilde y_{2k} = y_{k}$. Since most of our operations are performed on the augmented set, we will refer to this set of $2N$ samples as ``a training batch'' in the following part. The framework of UniCon is depicted in Fig. \ref{fig2}.

\subsection{Contrastive Loss}
Our proposed method is based on contrastive learning. As the most used contrastive loss, InfoNCE loss \cite{cpc} draws positive pairs close to each other while separating the negative ones. InfoNCE loss is defined in this form:

\begin{equation}\label{eq1}
L_{contrast} = - \frac{1}{2N}\sum_{i=1}^{2N}log \frac{exp(z_i\cdot z_{p(i)}/\tau)}{\sum_{k \neq i} exp(z_i \cdot z_k / \tau)},
\end{equation}
where $z_i = f(\widetilde x_i)$ represents the normalized deep embedding for each data point, $\tau$ is a temperature parameter, and $p(i)$ indicates a \emph{positive} for anchor $i$ while the rest indices are \emph{negatives}.

Considering that Eq. \ref{eq1} does not encode the label information, SupCon loss\cite{supcon} involves in-class samples into the positives:

\begin{equation} \label{eq2}
L_{sup} = \sum_{i=1}^{2N} \frac{-1}{|D_i|} \sum_{d \in D_i} log \frac{exp(z_i\cdot z_d/\tau)}{\sum_{k \neq i} exp(z_i \cdot z_k / \tau)},
\end{equation}
where $D_i \equiv \{k|k\in \{1,2,..,2N\}, k \neq i, \widetilde y_k=\widetilde y_i \}$ is a set of indices that refer to samples in the same class with i, and $|D_i|$ denotes the capacity of the set. Both two losses pay limited attention to negative pairs, simply recycling the non-positive sample pairs.

\subsection{Universum-style Mixup}
Motivated by universum learning, universum-style Mixup intends to provide a set of additional negatives to boost the performance of contrastive learning. It is assumed that by rejecting visual ambiguity, classes can be better separated with margins among them. Just like traditional Mixup method, Universum-style Mixup convexly combine each anchor sample $\widetilde x_i$ in a training batch, and its out-of-class negative $\widetilde x_{q(i)}$ to generate a universum negative $u_i$. Different from traditional Mixup strategy, in our approach the Mixup parameter $\lambda$ is set to a certain number rather than randomly sampled from Beta distribution. With this approach, we minimize the possibility of the universum data falling into the regions of target classes in the data space, thereby ensuring the negativity of Mixup-induced universum in a more principled way. The universum is acquired through the following process:
\begin{equation} \label{eq3}
u_i = \lambda \cdot \widetilde x_i + (1-\lambda) \cdot \widetilde x_{q(i)}, \quad i=1,2,..,2N,
\end{equation}
where q(i) is randomly chosen from $\cup_{k \neq i} D_k$ and $\lambda$ is the Mixup parameter. In the remainder of this paper, $u_i$ will be referred to as a ``g-negative'' and $\widetilde x_i$ will be referred to as its ``anchor''. Please note that universum-style Mixup does not mix the labels, and therefore the synthesized samples should belong to, if any, a generalized negative class. By doing so, our method completely drops the effect of label smoothing in Mixup \cite{mixup1}, in return earning a group of samples with hard labels. Furthermore, since the mixed data point is randomly sampled from out-of-class data, such Mixup can serve as an instance adaptive way \cite{adaptadv} to generate mildly adversarial data \cite{attacks} that contribute to the robustness of our model.

The expectation of $u_i$ is the mixture of $\widetilde x_i$ and all of its out-of-class negatives.
\begin{align}
E(u_i) &= \lambda \widetilde x_i + (1-\lambda)E(\widetilde x_{q(i)}) \\
&= \lambda \widetilde x_i +(1-\lambda) \frac{\sum_{k \notin D_i} \widetilde x_k}{2N-|D_i|}\label{ex}
\end{align}
Therefore, $u_i$ can be viewed as sampled from the classification boundaries in all directions with respect to $\widetilde x_i$. Compared to traditional Mixup data which are regarded as pseudo images with soft labels, universum-style Mixup images can be better interpreted as true data that belong to a generalized negative class so as to provide guidance for training with more reliability.

\subsection{Universum-inspired Supervised Contrastive Learning}
In this paper, our approach introduces a set of universum data ${u_k}_{k=1}^{2N}$ (which has been elaborated in Eq. \ref{eq3}) into the contrastive loss. The normalized encoded representation of $u_l$ is denoted as $zu_k=f(u_k)$. As Fig. \ref{fig2} shows, our proposed method intends to draw anchor samples close to the center of their class while pushing them from negatives. Here two solutions ($L_{add}$ and $L_{UniCon}$) are presented in the following parts.

\textbf{Universum data as additional negatives.}
A straightforward way of combining supervised contrastive learning and Mixup-induced universum is to use universum data as additional negatives. 
\begin{tiny}
\begin{equation} \label{eq4}
L_{add} = \sum_{i=1}^{2N} \frac{-1}{|D_i|} \sum_{d \in D_i} log \frac{exp(z_i\cdot z_d/\tau)}{\sum_{k \neq i} exp(z_i \cdot z_k / \tau)+\sum_{k=1}^{2N}exp(z_i \cdot zu_k / \tau)}
\end{equation}
\end{tiny}
$L_{add}$ generally adopts the original form of Eq. \ref{eq2}, yet further contrasting anchor samples with universum negatives. This loss function aims to use large amounts of universum negatives to alleviate the need for large amounts of negative samples in contrastive learning \cite{supcon, simclr}. However, as Table \ref{ablation} demonstrates, this loss function is not very effective on CIFAR-100 dataset. To justify such a result, here are two possible causes. On the one hand, it is deduced that the problem of ``manifold intrusion'' in Mixup (e.g. an image of number ``1'' and image of number ``4'' are mixed into a image that somewhat looks like number ``4'') may also appear in our universum-style Mixup, leading to poor results\cite{adversarial}. On the other hand, $L_{add}$ may overemphasize negatives, which produces undesirable disequilibrium.

These possible causes indicate that $L_{add}$ pays too much attention to universum negatives which possess too many noises, while the in-class positives are too clean to handle them. Therefore, comes up a natural idea that universum data should also be introduced into contrast with positives to maintain a balance between positives and negatives. Based on this idea, this paper proposes an entirely universum-based method.

\begin{figure}[htbp]
\centering
\includegraphics[width=0.4\textwidth]{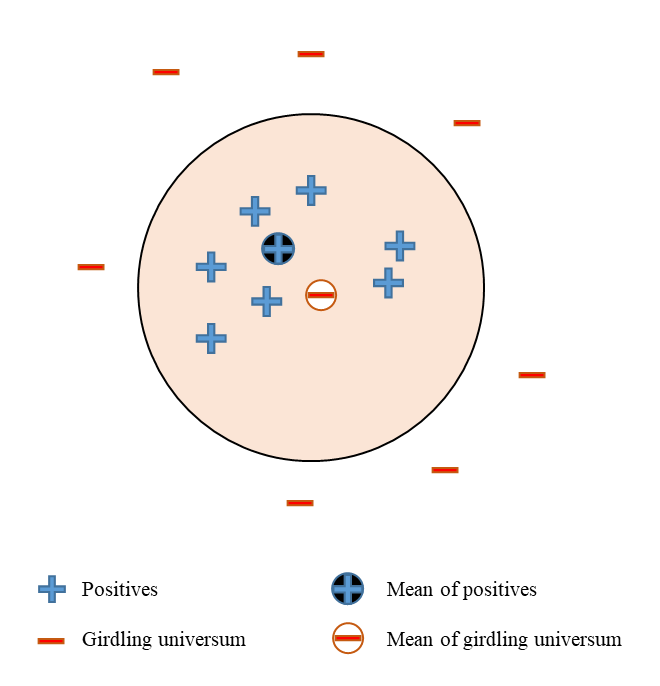}
\caption{The illustration of using the mean of universum data points for a class center. Here the universum data points are synthesized from the positives of the class in question. Since these universum data are expected to be distributed in the margin space girdling the in-class space, their mean may better describe the class especially when the positive samples are not evenly distributed in a minibatch.} \label{mean}
\end{figure}

\textbf{An entirely universum-based method.}
Here is the main loss function we use in this paper. This strategy is entirely based on universum data, both for contrast with negatives and derivation of class centers in the embedding space. It is worth noticing that since universum images are the equal mixture of two images from different classes, their features will naturally fall into the margin between two clusters. Eq. \ref{ex} has shown that these universum data are sampled from the decision boundaries in all directions. Therefore, as Fig. \ref{mean} illustrates, universum data points that are close to the cluster of $z_i$ are very likely to girdle the in-class space so their mean may serve as a better cluster center than the mean of in-class positives. According to these intuitions, our model pushes the anchors close to universum-based class centers rather than positives. The loss function is in the following form.

\begin{equation} \label{eq5}
L_{UniCon} = -\sum_{i=1}^{2N} log \frac{exp(z_i\cdot m_i/\tau)}{\sum_{k \neq i} exp(z_i \cdot zu_k / \tau)}
\end{equation}

%\begin{equation} \label{eq5}
%L_{UniCon} = \sum_{i=1}^{2N} \frac{-1}{|D_i|} \sum_{d \in D_i} log \frac{exp(z_i\cdot zu_d/\tau)}{\sum_{k \neq i} exp(z_i \cdot zu_k / \tau)}
%\end{equation}

%To differentiate Eq. \ref{eq5} from a na\"ive combination of Mixup-induced universum and Eq. \ref{eq2}, $L_{UniCon}$ is further derived into the following form.

%\begin{small}
%\begin{align}
%    L_{UniCon} &=\sum_{i=1}^{2N} \frac{-1}{|D_i|} \left[ \frac{z_i}{\tau}\cdot \sum_{d \in D_i}zu_d - \sum_{d \in D_i}log\sum_{k \neq i}exp(z_i\cdot zu_k/\tau)\right] \\
%    &=\sum_{i=1}^{2N}\left[-\frac{z_i}{\tau}\cdot m_i+log\sum_{k \neq i} exp(z_i \cdot zu_k / \tau)\right]\label{newloss}
%\end{align}
%\end{small}
\noindent where $m_i=(\sum_{d \in D_i}zu_d)/|D_i|$ is the mean of the representations of universum data points around the cluster of $z_i$. 
According to Eq. \ref{eq5}, $L_{UniCon}$ drives in-class data points close to class center $m_i$, which is derived from universum data. 
Meanwhile, $L_{UniCon}$ only adopts universum data as negatives, dropping out negatives in the conventional sense, which further improves model robustness. Still, it should be admitted that this strategy is coarse and primary, yet the experimental results show that it is especially effective.

Table. \ref{ablation} empirically demonstrates that $L_{UniCon}$ works better than $L_{add}$. The performance of $L_{add}$ is even worse than the loss without the extra universum negatives, which implies that an entirely universum-based framework is crucial for utilizing the universum data. Based on these findings, we deduce that our method generalizes better to the test set for the following reasons:

\textbf{Noise injection.} In the aforementioned situation, our method injects noises to the training data (e.g. anchors in class ``4'' regard number``4'' synthesized by ``1'' and ``4'' as a negative sample). On the one hand, such technique is widely used in adversarial training as well as contrastive learning to learn a more robust model \cite{adv-fellow,adversarial,clusteralone}. On the other hand, since Mixup-induced universum are used in both contrast with negatives and class centers, these two kinds of contrast are in a restrictive relation with each other. Noises in universum negatives can help derive a more accurate class center, and vice versa.

\textbf{A different approach of contrast} Our method does not directly contrast anchors with conventional out-of-class negatives in \cite{supcon}. However, UniCon still uses universum data as negatives, which differentiates itself from absolutely contrast-free methods like \cite{byol}. By contrasting with universum negatives and benefiting from their data diversity, UniCon not only avoids contrastive models' dependency on large batch sizes, but also allows a balanced network design easier to optimize.

\subsection{Unsupervised Universum-inspired Contrastive Model}
Our method could also be applied to the unsupervised setting. Considering the specific situation of the task, the loss function is adapted to the following form: 
\begin{equation}
    L_{Un-Uni} = L_{contrast} + L_{Uni}
\end{equation}
Here $L_{contrast}$ is the SimCLR loss which can be computed through Eq.\ref{eq1}. In this setting, we sum up $L_{Uni}$ and an original contrastive loss for the ultimate Un-Uni loss function. The unsupervised version of $L_{UniCon}$ is derived by simply setting the labels of a minibatch to [0, 1, .., N-1]. In this way, the loss function becomes the following form.
\begin{equation} \label{unsupervised}
L_{Uni} = -\sum_{i=1}^{2N} log \frac{exp[z_i\cdot \frac{1}{2}(zu_i+zu_{p(i)})/\tau]}{\sum_{k \neq i} exp(z_i \cdot zu_k / \tau)}
\end{equation}
where p(i) represents the index of the only corresponding positive for each anchor (i.e. the counterpart in a [i, i+N] pair (i=1, 2, .., N)). Similarly, the anchors are drawn close to an instance center derived from the mean of related universum pseudo-positives. Please note that the term $L_{contrast}$ is necessary since each instance center is derived from only two data points and therefore could be easily influenced by noises.

\subsection{Theoretical Analysis} \label{theory}
With analysis on gradients, we intend to show that UniCon loss not only has the effect of hard negative mining, but also helps maintain large margins among different classes. Since the theoretical foundations of the contrastive framework abounds, we would focus on the effectiveness of our universum images.

\textbf{Hard negative mining.} Following \cite{supcon}, we calculate the gradients of $L_{UniCon, i}$ with respective to $z_i$. The gradient can be written in the following form:

\begin{equation}\label{gradient1}
    \frac{\partial L_{UniCon,i}}{\partial z_i} =
    \frac{1}{\tau}\left[-m_i+\sum_{k \neq i}zu_kPU_k+G\right]
\end{equation}
where we define,
\begin{equation}
PU_k = \frac{exp(z_i \cdot zu_k / \tau)}{\sum_{j\neq i}exp(z_i \cdot zu_j / \tau)}
\end{equation}
\begin{equation}
G = \frac{z_i \sum_{k\neq i} exp(z_i \cdot zu_k / \tau)\frac{\partial zu_k}{\partial z_i}}{ \sum_{k \neq i}exp(z_i \cdot zu_k / \tau)}
\end{equation}

The gradient can be divided into three parts: the representation of the class center, gradient of universum negatives, and universum gradient $G$. Obviously, the optimization process is always influenced by the class center $m_i$. In line with \cite{supcon}, UniCon loss also inherits the inner ability of hard negative mining. Here we show it with gradient of universum negatives. When a universum negative representation $zu_k$ is hard, $z_i \cdot zu_k \approx 1$, otherwise $z_i \cdot zu_k \approx 0$. Apparently, the harder $zu_k$ is, the larger $P_n$ becomes, and therewith the greater its influence towards the optimization. The details can be found in the supplementary material.

\textbf{Large margin maintenance.} By further calculating the gradients of $zu_k$ with respect to $\widetilde x_i$, we derive the following form of $G$:
\begin{align}  
  G &= \frac{(1-\lambda)z_i}{f^{'}(\widetilde x_i)} \sum_{k \in Q_i}f^{'}(u_k)PU_k
\end{align}
where we denote $Q_i = \left \{k|q(k) = i\right\}$ and $f$ is the deep encoder. It is worth noticing that $G$ increases with $f^{'}(u_k)$. Since $u_k$ is the mean of two images of different classes, $f(u_k)$ would naturally fall into the margin between these two clusters. We conjecture that when $f^{'}(u_k)$ is large, the gradients on the universum data points are sharp and changing quickly due to the reason that the margins among different classes are narrow in the deep embedding space. In contrast, small $f^{'}(u_k)$ suggests gentle gradients and wide margins. Our model is expected to benefit from $G$ in the former situation and draw clearer decision boundaries by converging to the latter situation.

\begin{table}[H]
\setlength{\tabcolsep}{1.2mm}
\renewcommand\arraystretch{1.5}
\caption{Dataset settings.}\label{tab4}
\centering
\begin{tabular}{cccc}
\hline
Dataset&Images&Classes& Input Size\\
\hline
CIFAR-10&60,000&10&$32\times 32$\\
CIFAR-100&60,000&100&$32\times 32$\\
TinyImageNet&100,000&200&$32\times 32$\\
ImageNet-100&130,000&100&$64 \times 64$\\
\hline
\end{tabular}
\end{table}

\section{Experiments}
\subsection{Setup}
We evaluate our model on several widely used benchmarks including CIFAR-10, CIFAR-100 \cite{cifar}, TinyImageNet \cite{tinyimagenet}, and a 100-category subset (ImageNet-100) of ImageNet \cite{russakovsky2015imagenet}. For ImageNet-100, we choose the first 100 categories with smallest category numbers in 1k classes. Detailed information of dataset settings can be viewed in Table. \ref{tab4}. Here input size refers to the transformed size of neural network input. Without special statement, the encoder network is trained for 1000 epochs with a batch size of 256. As for hyperparameters, temperature $\tau$ and Mixup parameter $\lambda$ are respectively fixed to 0.1 and 0.5. We set the learning rate to 0.05 with 10 epochs of warm-up. As the purpose of this paper is to show how universum improves contrastive learning rather than to explore the effects of different augmentation techniques on our model, we empirically use a set of augmentations that was chosen by \cite{supcon} through AutoAugment \cite{autoaugment}. The details of our augmentations are written in the supplemental material. In the evaluation period, a classifier of batch size 512 is trained for 100 epochs with the deep representations extracted by the encoder while the encoder itself is frozen. Compared to the prior version, we separate the representations and the contrastive features and implement classification on the learned representations, which further improves the results. On both stages, we use SGD optimizer with cosine annealing for weight decay. Our experiments are implemented in PyTorch framework on at most four Nvidia Tesla V100 GPUs in an online computing center.

\begin{table}[]
    \centering
    \renewcommand\arraystretch{1.5}
    \caption{Classification results (\%) on Imagenet-100 with ResNet-18 as the backbone.}
    \begin{tabular}{|c|c|c|c|c|}
    \hline
         ImageNet-100 & Xent & SimCLR & SupCon & UniCon  \\
         \hline
         Top-1 Accuracy & 72.9 & 50.8& 69.8 & 77.3 \\
    \hline
    \end{tabular}
    
    \label{imagenet}
\end{table}

\begin{table*}[htbp]
\setlength{\tabcolsep}{1.5mm}
\renewcommand\arraystretch{1.2}
\caption{Top-1 classification accuracy (in percentage \%) on various datasets. We compare our model (UniCon) with a deep classifier using cross-entropy loss, SimCLR\cite{simclr}, and SupCon\cite{supcon}. We re-implement the results for baseline models while showing the published numbers of SupCon. We use \textbf{bold} to indicate the best results, and \underline{underline} the second best ones. Also please note that the batch size of our model is only 256, which is much smaller than that of the baseline models.}\label{tab1}
\centering
\begin{tabular}{cccccc}
\hline
Method &  Architecture & Batch size&CIFAR-10 & CIFAR-100 & TinyImageNet\\
\hline
Cross-Entropy & ResNet-50 & 1024&94.6 &77.2 &58.3\\
SimCLR &  ResNet-50 & 1024& 91.8 & 68.4&51.2\\
SupCon(baseline) & ResNet-50&1024&96.0&76.5& - \\
SupCon(our impl.) &  ResNet-50 & 1024&95.9 & 75.4 & 58.3\\
\hline
\multirow{2}{*}{UniCon(ours)} & ResNet-18 & \emph{\textbf{256}}& \underline{96.4} & \underline{79.2}& \underline{59.3}\\
 & ResNet-50 & \emph{\textbf{256}}&\textbf{97.4} & \textbf{81.7}& \textbf{65.0}\\
\hline
\end{tabular}
\end{table*}

\subsection{Classification Accuracy}
We compare UniCon with a cross-entropy classifier, SimCLR \cite{simclr}, and SupCon \cite{supcon} on their top-1 accuracy on CIFAR-10, CIFAR-100 and TinyImageNet. Although these methods have all be proposed for a few years, so far they are the mainstream methods for fully-supervised learning. Follow-up methods either focus on a specific application scenario or adapt the aforementioned models to other settings, failing to propose a better model on fully-supervised learning. Therefore, we still adopt these three old but effective models as our baselines. We re-implement all the baseline models while also showing the published numbers of SupCon. As is shown in Table. \ref{tab1}, UniCon outperforms other models on all datasets, while adopting smaller batch sizes and encoder backbones. Our model achieves 97.4\%, and 81.7\% on CIFAR-10 and CIFAR-100, respectively, which surpasses the state of art (published numbers) by a significant margin of 1.4\% and 5.2\% with only one fourth the batch size. Even with a backbone of ResNet-18 and batch size 256, UniCon outperforms its counterparts with ResNet-50 and batch size 1024. UniCon also achieves 65.0\% top-1 accuracy on TinyImageNet. Please note that we input images into the neural network as $32\times 32$ patches, which is way smaller than the input sizes (e.g. $224 \times 224$) of other models \cite{mixco} that report better performance of cross-entropy classifiers. In this sense, our performance gain over the cross-entropy classifier is also significant on TinyImageNet.

To further verify the model performance, we also evaluate UniCon, Xent, SimCLR and SupCon on ImageNet-100 with ResNet-18 being the backbones. The training batchsize is 256 for UniCon and 1024 for others. As is shown in Table \ref{imagenet}, UniCon greatly outperforms other models by 7.5\%, which proves UniCon's ability of handling large datasets in a nutshell. 

\begin{table*}[htbp]
\setlength{\tabcolsep}{2mm}
\renewcommand\arraystretch{1.5}
\caption{Comparison with other models assisted with Mixup/CutMix/Augment. "Augment" here refers to the same augmentation techniques we exert on our model. We implement all models on CIFAR-100 with ResNet-18 as their backbones. In this comparison, UniCon outperforms the cross-entropy classifier even when the CE classifier is boosted by both augmentations and MixUp. \IEEEauthorrefmark{2}: Details of our implementation can be found in the supplementary material.}\label{tab5}
\centering
\begin{tabular}{lcccc}
\hline
Method  & Augment & Mixture &  Top-1 & Top-5\\
\hline
Cross-Entropy  &  & -  &74.9 & 90.5\\
Cross-Entropy + Augment  &  \checkmark & -  & 76.8 & 90.9\\
Cross-Entropy + Mixup  &  & MixUp &&\\
Cross-Entropy + CutMix  &  & CutMix & & \\
Cross-Entropy + Augment + Mixup  & \checkmark & MixUp & 78.1& 92.6\\
Cross-Entropy + Augment + Cutmix  & \checkmark & CutMix &  78.5& 93.1\\
\hline
SupCon & \checkmark & - &  72.3 & 90.7\\
SupCon + Un-Mix \IEEEauthorrefmark{2} & \checkmark & MixUp  & 75.1 & 92.7\\
SupCon + Un-Mix (universum-style) & \checkmark & MixUp  & 78.4 & 94.5\\
\hline
UniCon(ours) + CutMix & \checkmark & CutMix  & 72.9 & 91.2\\
UniCon(ours) & \checkmark & MixUp  & \textbf{79.2}& \textbf{94.6}\\
\hline
\end{tabular}
\end{table*}

\begin{table*}[htbp]
\caption{Universum-style Mixup improves the robustness of supervised contrastive learning. (\textbf{Left:}) Here the robustness of various supervised models are measured with Error Rate on the clean dataset (Err.), Mean Corruption Error (mCE) and Relative Mean Corruption Error (rel. mCE) (lower is better for all metrics) on CIFAR-100-C dataset and TinyImageNet-C dataset. All models are only trained on the clean datasets. (\textbf{Right:}) Top-1 Accuracy with different corruption severity (higher is better).}\label{tab10}
\subfigure[]{
\begin{minipage}[]{0.6\textwidth}
\setlength{\tabcolsep}{1.5mm}
\renewcommand\arraystretch{1.5}

\centering
\begin{tabular}{l|c|ccc|ccc}
\hline
\multirow{2}{*}{Model} & \multirow{2}{*}{Architecture}  & \multicolumn{3}{c|}{CIFAR-100-C} & \multicolumn{3}{c}{TinyImageNet-C} \\
\cline{3-8}
& &Err.($\downarrow$) & mCE($\downarrow$) & rel.mCE($\downarrow$)&Err.($\downarrow$) & mCE($\downarrow$) & rel.mCE($\downarrow$) \\
\hline
\multirow{2}{*}{Cross Entropy} & AlexNet& 42.9 & 100.0& 100.0& 62.1& 100.0&100.0 \\
%\cline{2-8}
& ResNet-50&22.8 & 79.0 & 83.7 &41.7 &89.9 &86.5 \\
\hline
SupCon & ResNet-50 &24.6 &84.3 &\textbf{55.5} &41.7 &84.8 &80.0 \\
\hline
\multirow{2}{*}{UniCon(ours)}& ResNet-18&20.8 &73.4 &78.0 &40.7 &83.2 &77.8 \\
%\cline{2-8}
& ResNet-50&\textbf{18.3} &\textbf{70.6} &75.7 &\textbf{35.0} &\textbf{76.4} &\textbf{68.8} \\

\hline
\end{tabular}
\end{minipage}
}
\subfigure[]{
\begin{minipage}[]{0.4\textwidth}

    \centering
    \includegraphics[width=0.8\textwidth]{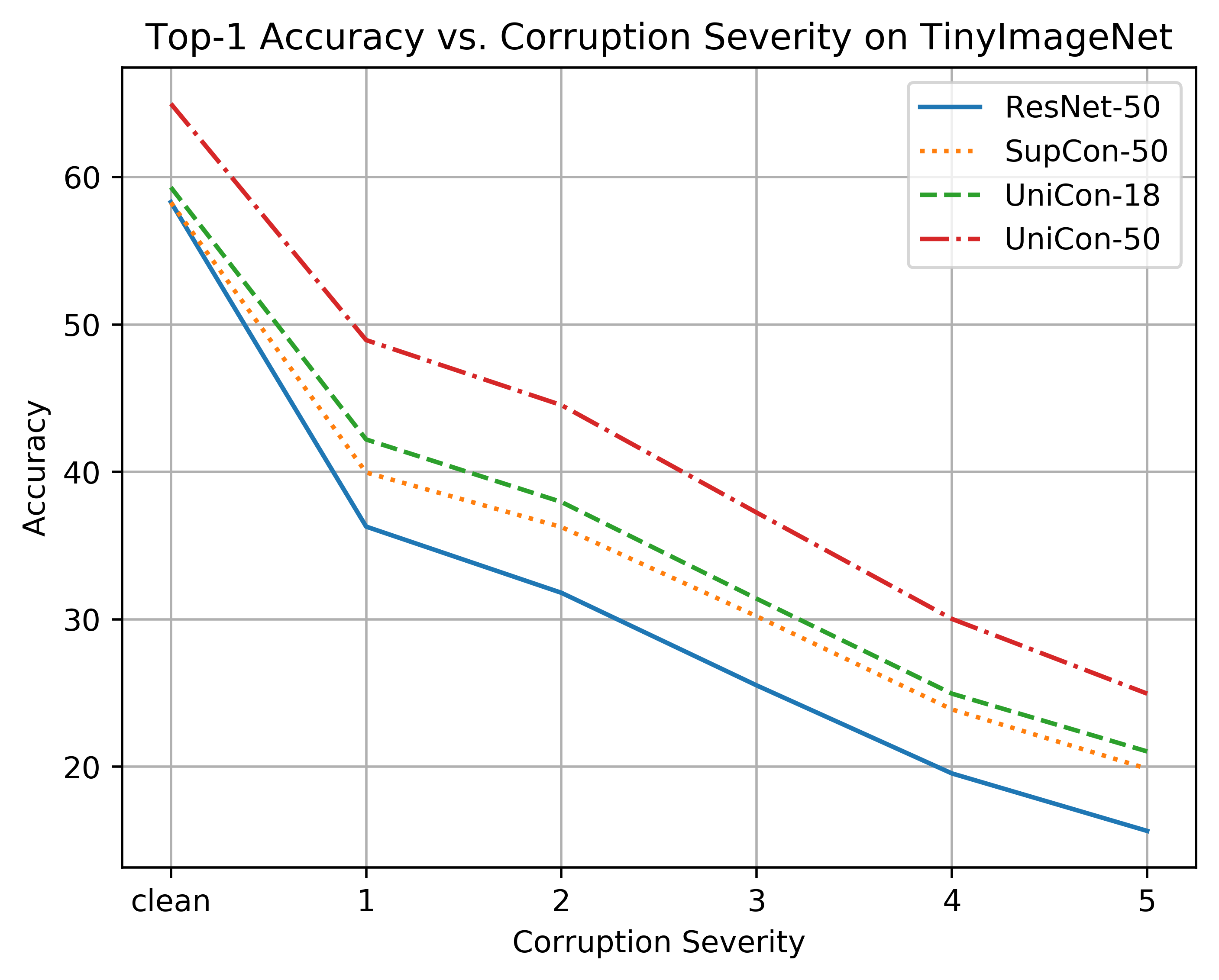}
    \label{fig:corruption}
\end{minipage}
}
\end{table*}
In Table. \ref{tab5} UniCon is compared with Xent and SupCon assisted with MixUp, CutMix and Augment. Here "Augment" refers to exactly same set of augmentations we use for our model. This augmentation method is also adopted by SupCon as the best augmentation strategy chosen by AutoAugment. When the cross-entropy classifiers are implemented with Augment, they not only use Augment to modify the input images but also double the size of the training data in the same way as contrastive models do. We also incorporate MixUp and CutMix with Xent, SupCon and UniCon. For Xent, we combine the baseline algorithm of MixUp and CutMix with our own implementations. For SupCon, since a naïve application of Mixup may result into unclear labels hard to handle in contrastive learning, we use the idea of Un-Mix \cite{unmix} to realize a non-universum Mixup-boosted supervised contrastive model. The details of this implementation can be found in the supplementary material. We did not implement the combination of SupCon and MoCHi\cite{hardmix} for the reason that our comparisons do not include memory-bank-based methods. All models are implemented with ResNet-18 as their backbones, while the batch size varies. We empirically find that for models with Mixup or CutMix perform better with a smaller batch size of 256, while other models benefit from a larger batch size of 1024. As is shown in the table, Xent boosted with Augment and CutMix greatly outperforms SupCon, while out model still surpasses it by 0.7\%. We also attempt to replace Mixup with CutMix in UniCon and the results show that Mixup performs better than CutMix in combination with UniCon. We deduce there are two reasons: (i) in CutMix, it is hard to set the Mixup parameter $\lambda$ to exact 0.5 (in our implementation, we accept $ 0.45 < \lambda < 0.55$), while $\lambda=0.5$ is crucial to UniCon as is shown in Section \ref{mixup}; (ii) our model may benefit from "the confusion when choosing cues for recognition" in Mixup as \cite{cutmix} argues.

\begin{figure*}[htbp]
    
	\centering
	\subfigure[Xent without universum negatives]{
		\begin{minipage}[b]{0.3\textwidth}
			\includegraphics[width=1\textwidth]{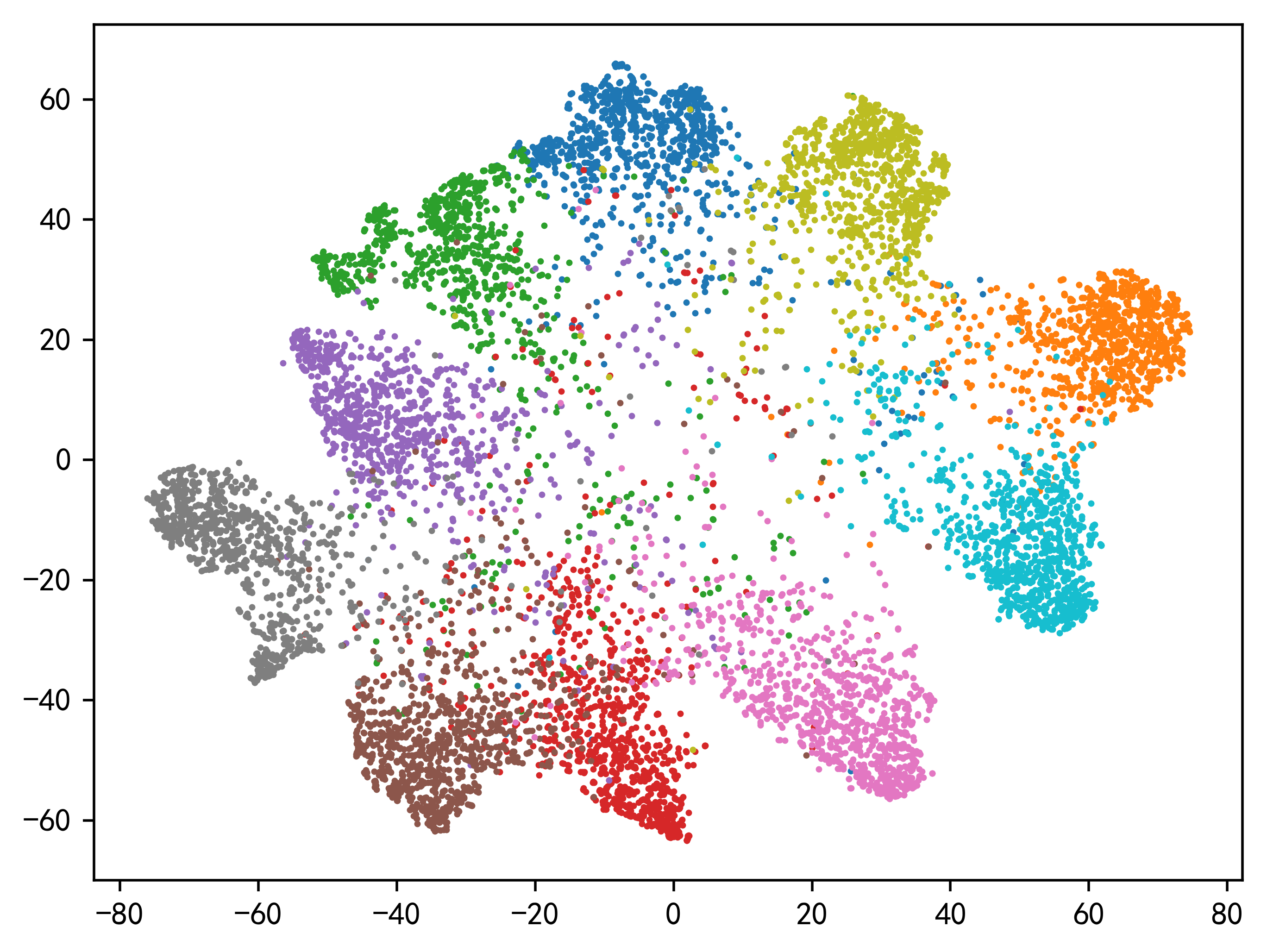} 
		\end{minipage}
	}
    	\subfigure[SupCon without universum negatives]{
    		\begin{minipage}[b]{0.3\textwidth}
   		 	\includegraphics[width=1\textwidth]{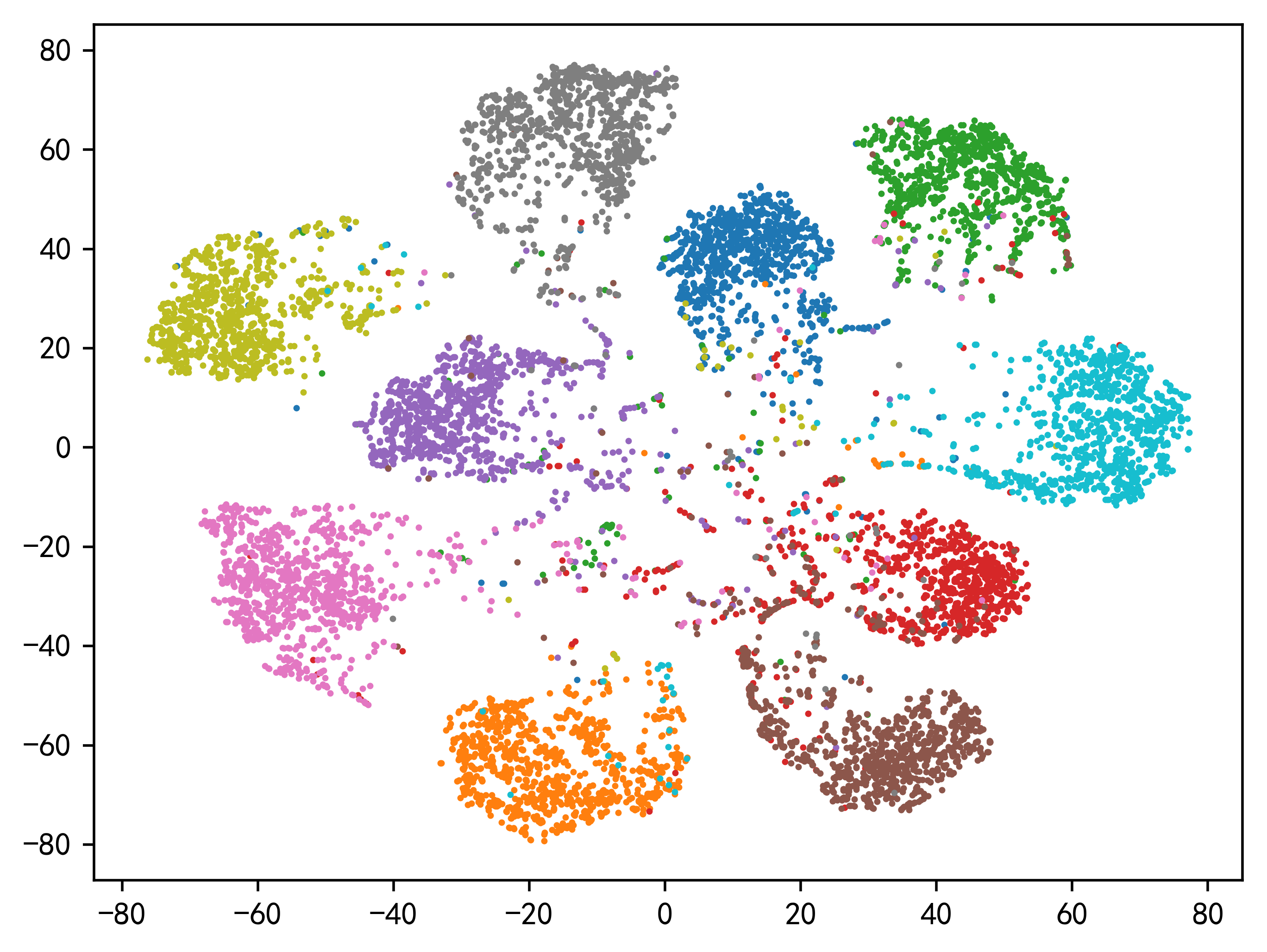}
    		\end{minipage}
    }
	    \subfigure[UniCon(ours) without universum negatives]{
		\begin{minipage}[b]{0.3\textwidth}
			\includegraphics[width=1\textwidth]{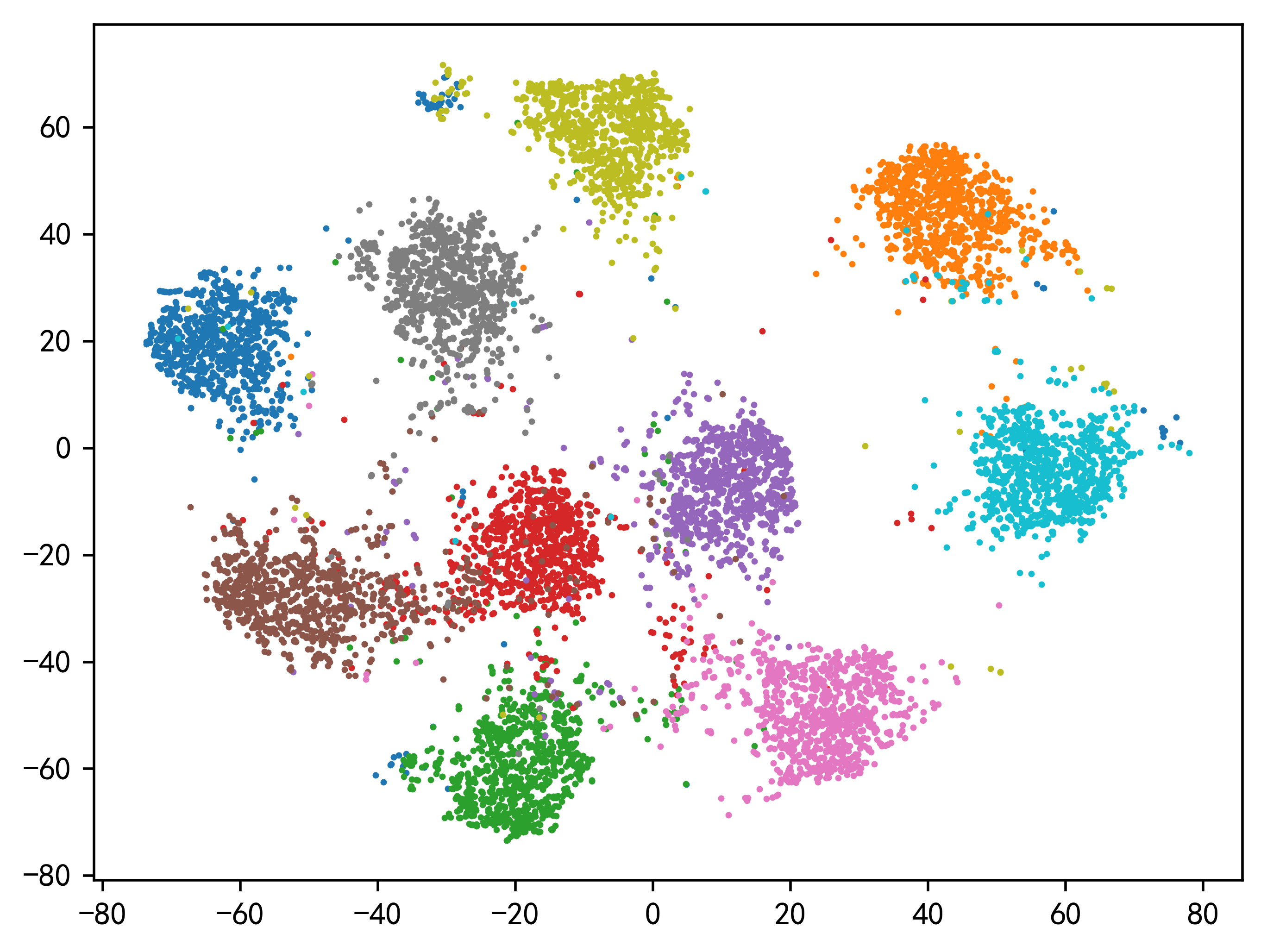} 
		\end{minipage}
	    }
    	\subfigure[Xent with universum negatives]{
    		\begin{minipage}[b]{0.3\textwidth}
		 	\includegraphics[width=1\textwidth]{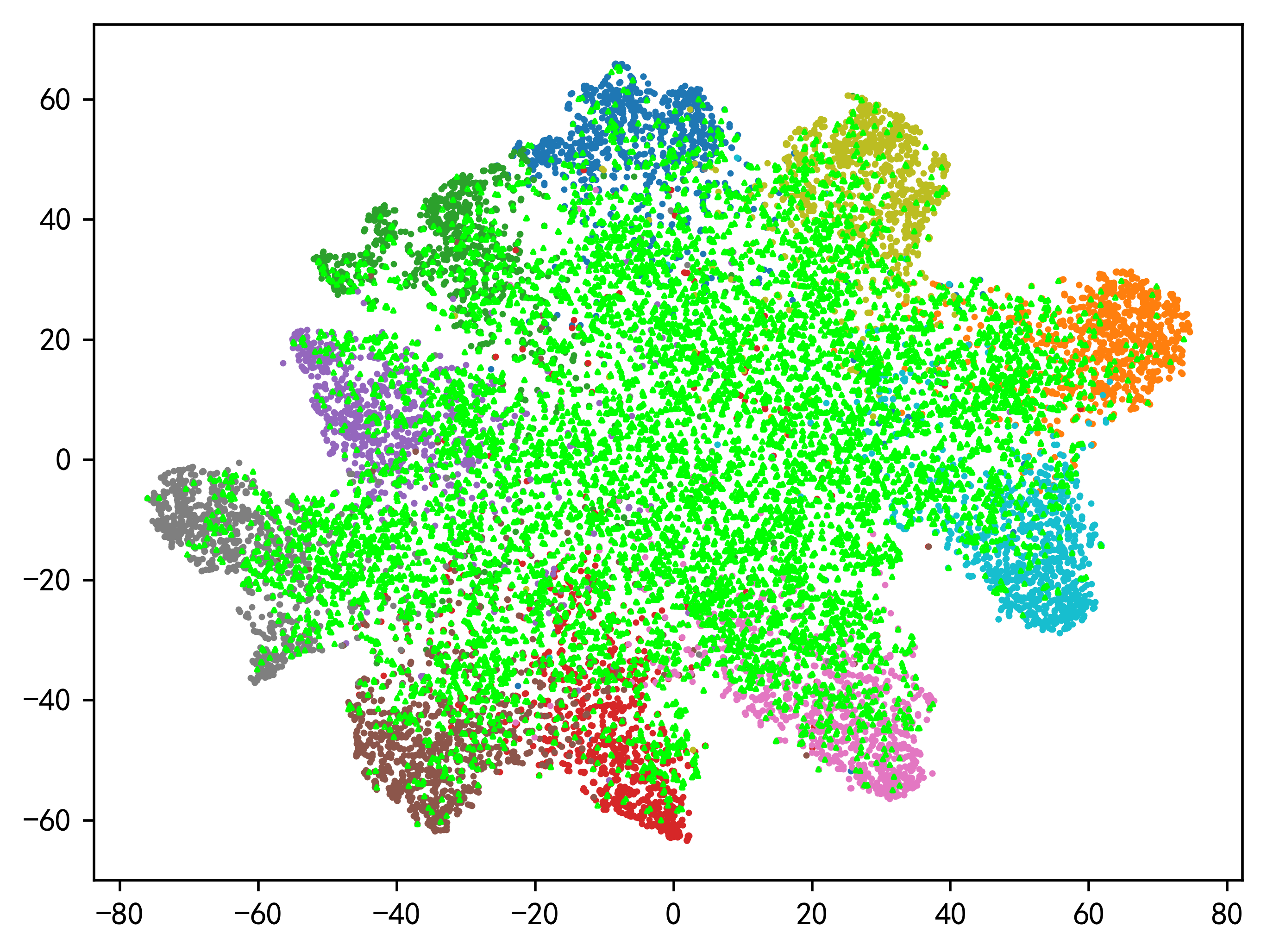}
    		\end{minipage}
    	}
    	\subfigure[SupCon with universum negatives]{
    		\begin{minipage}[b]{0.3\textwidth}
		 	\includegraphics[width=1\textwidth]{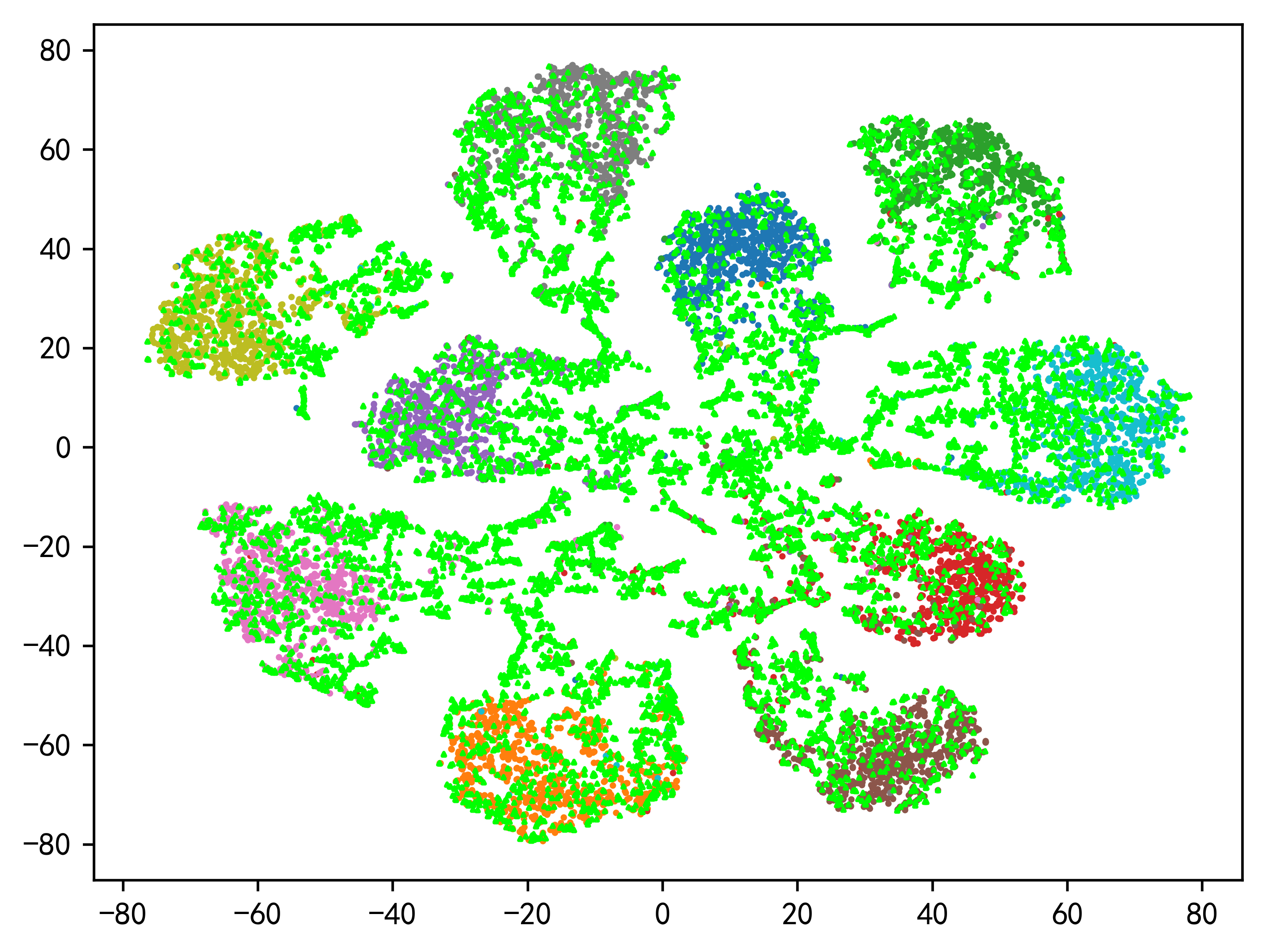}
    		\end{minipage}
    	}
    	\subfigure[UniCon(ours) with universum negatives]{
    		\begin{minipage}[b]{0.3\textwidth}
		 	\includegraphics[width=1\textwidth]{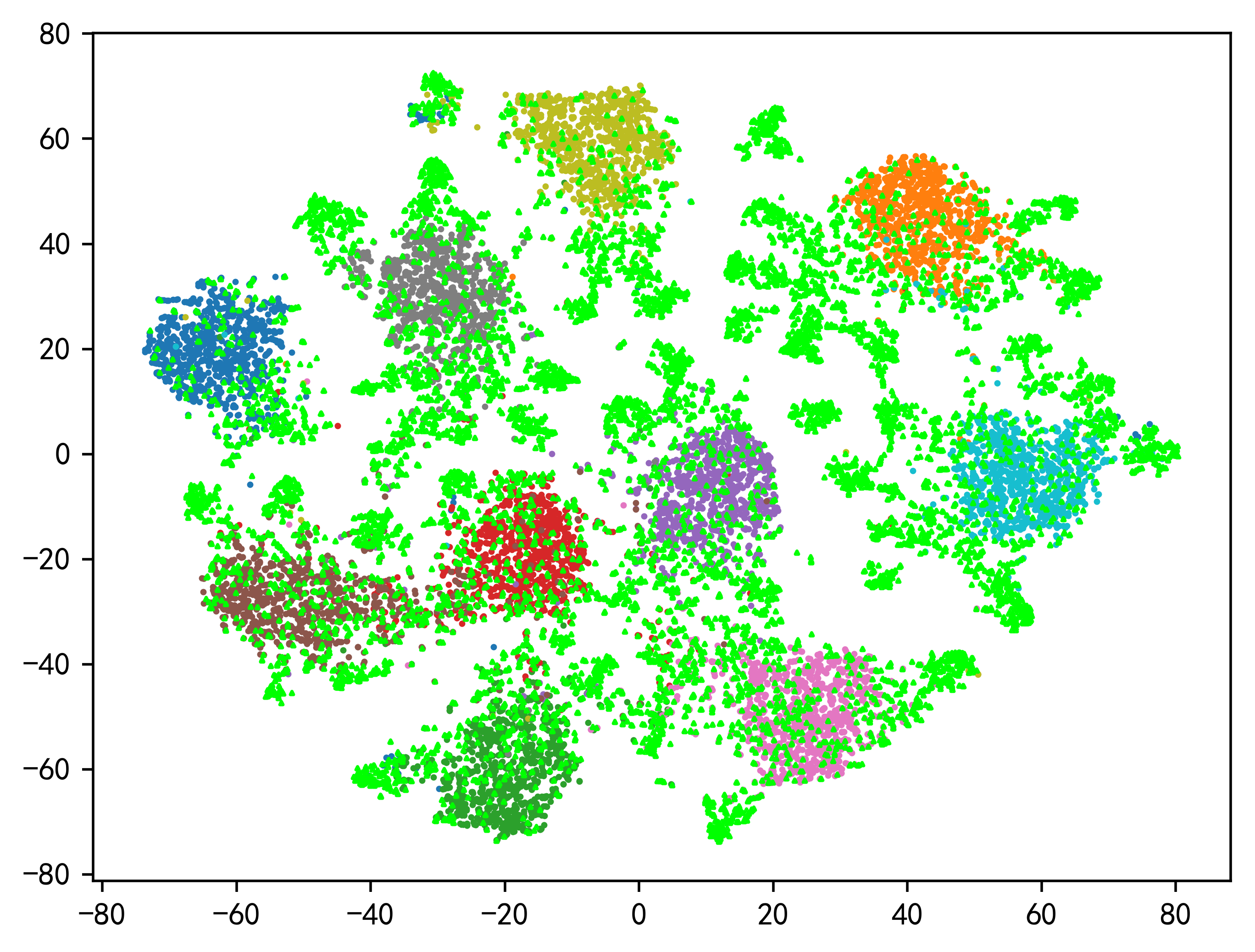}
    		\end{minipage}
    	}
	\caption{T-SNE visualizations of Xent, SupCon and UniCon with and without universum negatives on CIFAR-10. Specially the embeddings of universum data are colored lime. In the embedding space of both Xent and SupCon, manually synthesized universum negatives are mostly distributed nearby images, while UniCon places a large number universum negatives in the margin among different clusters. }\label{fig4}
\end{figure*}

\begin{figure*}[htbp]
    
	\centering
	\subfigure[]{
		\begin{minipage}[b]{0.23\textwidth}
			\includegraphics[width=1\textwidth]{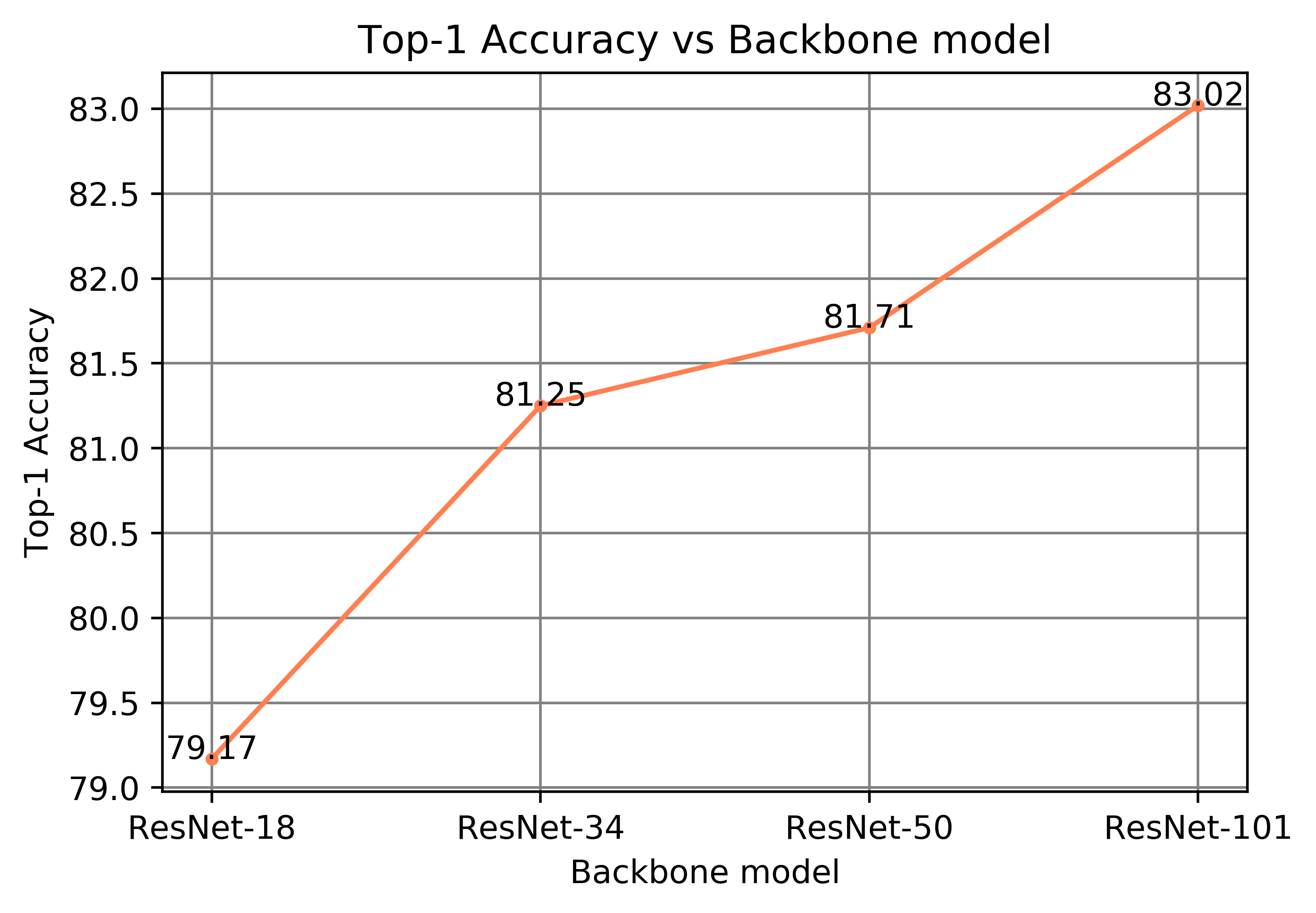} 
		\end{minipage}
	}
    	\subfigure[]{
    		\begin{minipage}[b]{0.23\textwidth}
   		 	\includegraphics[width=1\textwidth]{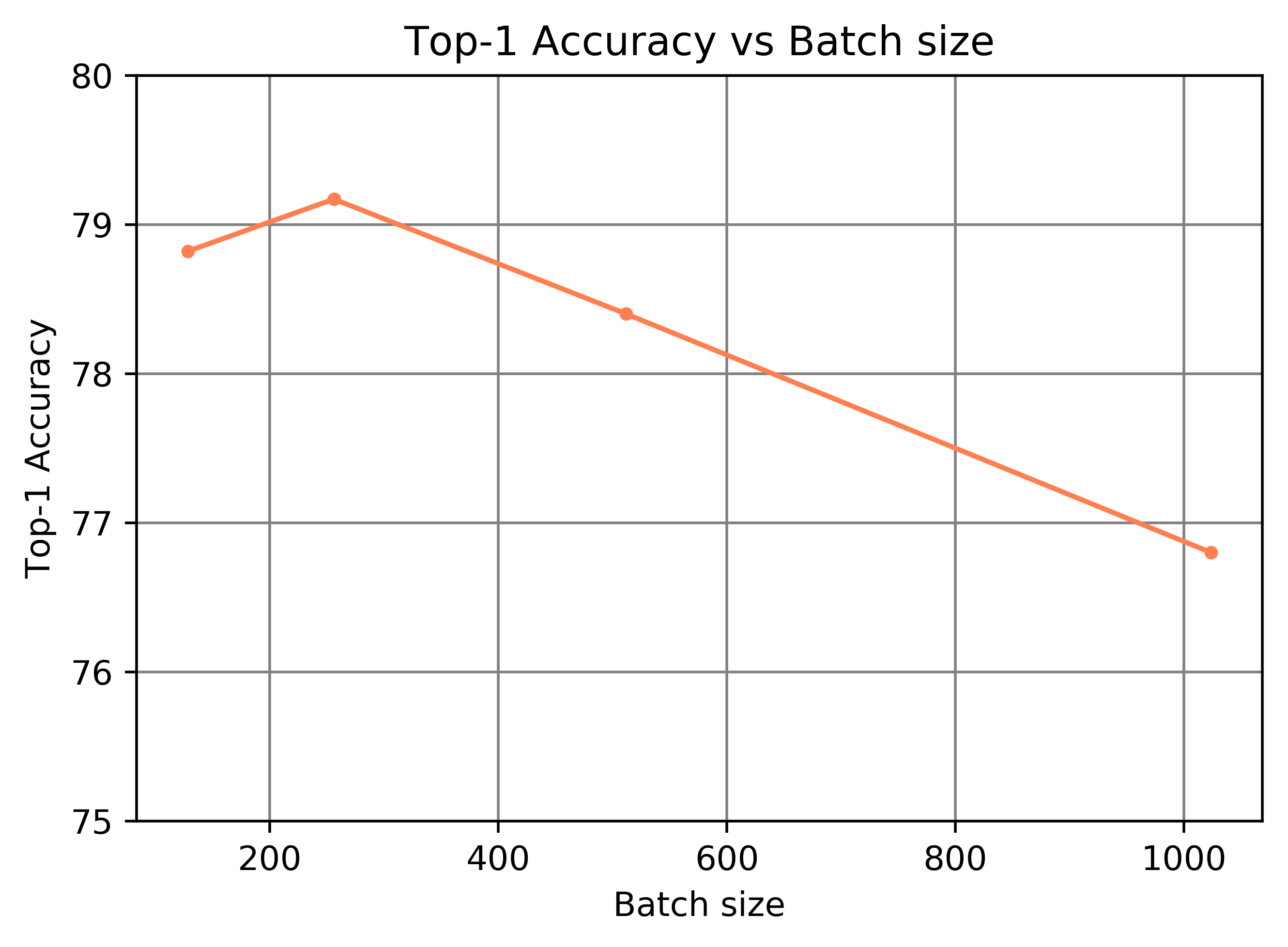}
    		\end{minipage}
    }
	    \subfigure[]{
		\begin{minipage}[b]{0.23\textwidth}
			\includegraphics[width=1\textwidth]{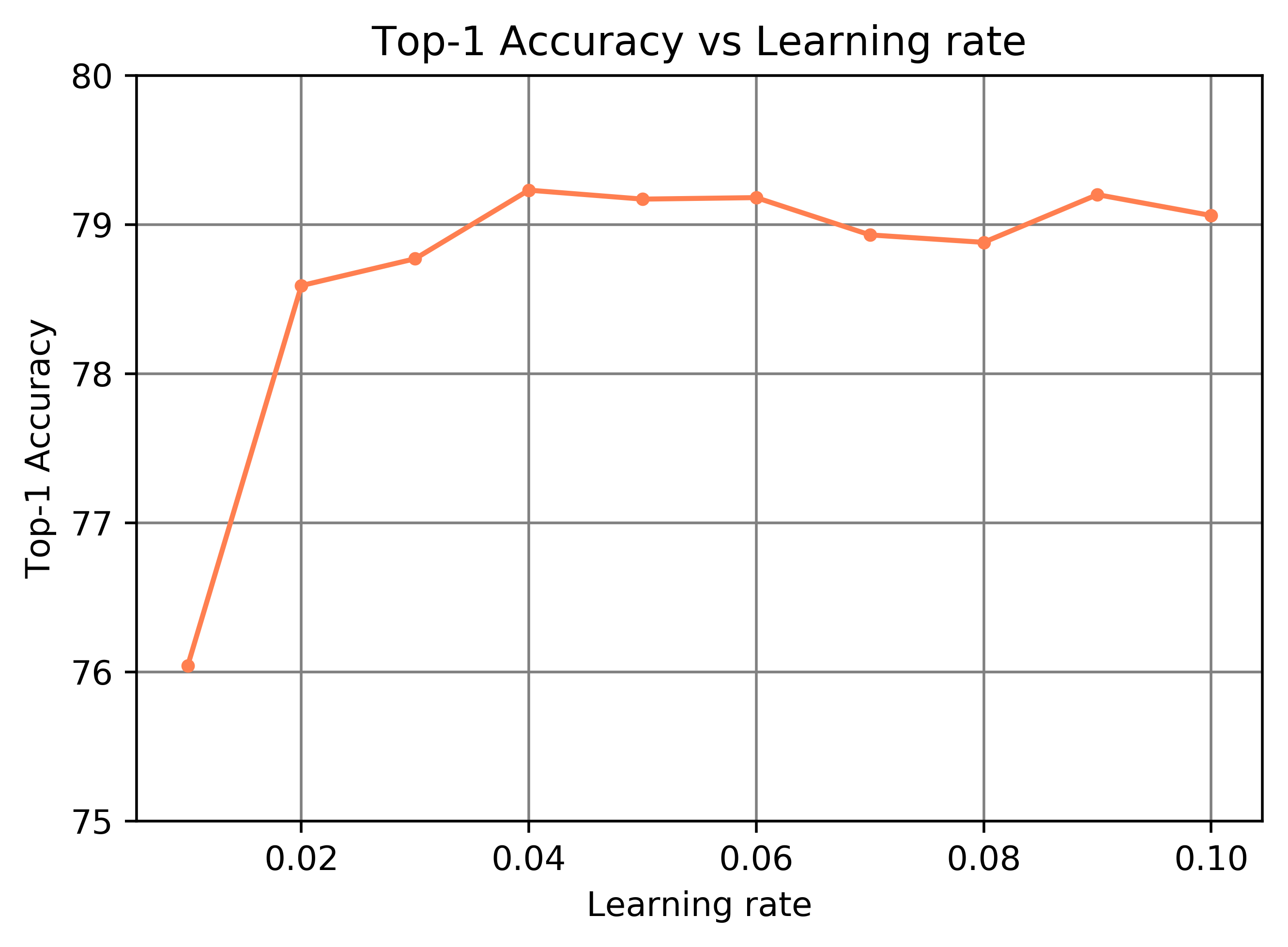} 
		\end{minipage}
	    }
    	\subfigure[]{
    		\begin{minipage}[b]{0.23\textwidth}
		 	\includegraphics[width=1\textwidth]{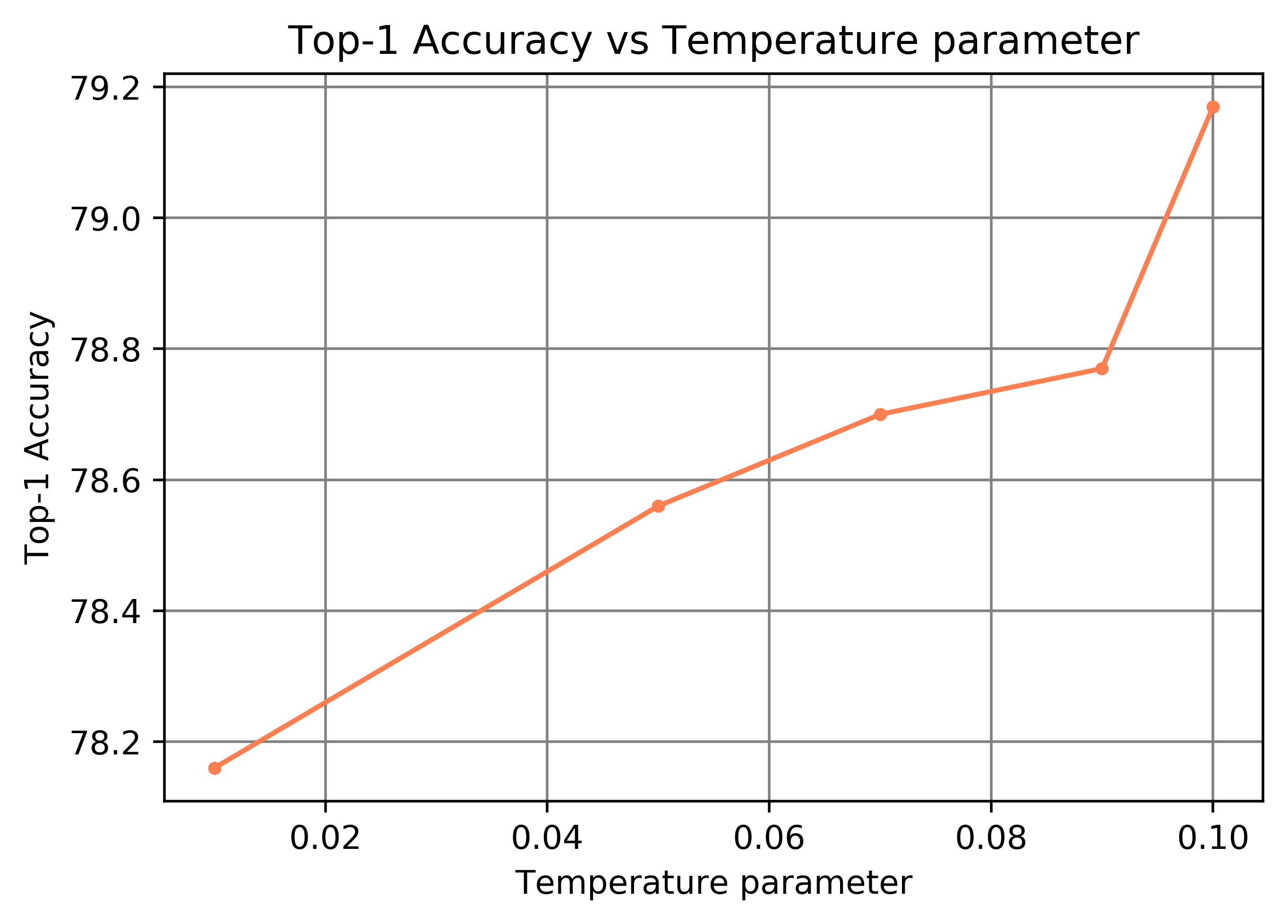}
    		\end{minipage}
    	}
	\caption{Top-1 Accuracy of UniCon with varying backbones, batch sizes, learning rates and temperature parameter. The experiments are conducted on CIFAR-100, and except for the backbone analysis, Resnet-18 is adopted for model encoders. }\label{fig3}
\end{figure*}

\subsection{Visualization Analysis}
We use t-SNE \cite{tsne} with the features extracted from Xent, SupCon and UniCon on CIFAR-10 test data to acquire 2-dimension visualizations. The first row of Fig. \ref{fig4} shows the distributions of the features of training data points, while the second row adds the distributions of the features of manually synthesized universum negatives. For the visualization, we specially synthesize 10,000 universum data from the test data with Mixup parameter 0.5 and use them for all three models. Compared with Xent and SupCon, our model better separates different clusters with large margins among them in the embedding space. Please note that the distribution of universum negatives with three models varies: for Xent, universum negatives disperse all over the space; for SupCon, they distribute alongside the original data points; for UniCon, a large number of universum data are placed in the margins among different clusters. As is assumed in Section \ref{sec:introduction}, universum negatives are hard in that Xent and SupCon tend to assign them to a known class rather than place them in the margins. Although the universum negatives are mixed from two images of different classes, in Fig. \ref{fig4}(e) they are apparently assigned to \textbf{one} of the two classes, which reveals an undesirable twist of the manifold space that the decision boundaries might be skewed to one of the classes. By driving universum data into the margin space, different classes are better separated and clearer decision boundaries are drawn.

\subsection{Robustness}
One of our assumptions is that like traditional Mixup method, Universum-style Mixup should also make our model more robust. CIFAR-100-C dataset and TinyImageNet dataset are two neural network robustness benchmark datasets derived from CIFAR-100 and TinyImageNet with deliberate corruptions including Gaussian noise, frost, elastic transform, jpeg compression, etc., the severity of which varies from 1 to 5 \cite{corruption}. The performance of AlexNet, Xent-50, SupCon-50, UniCon-18 and UniCon-50 are measured in the aforementioned approach. Except for AlexNet, the weights of all other models are the same with the ones reported in Table. \ref{tab1}. AlexNet is specially implemented for normalizing mCE and relative mCE as is required by \cite{corruption}. Since the classical implementation of AlexNet cannot handle images as small as $32\times32$, we slightly modify its convolutional kernels to train it on benchmark datasets. All models are only trained on the clean datasets. 

As is shown in Table \ref{tab10}(left), UniCon's mCE is the lowest on both corrupt datasets, and its relative mCE outperforms other models on TinyImageNet-C. Although UniCon's relative mCE on CIFAR-100-C is higher than other models, we deduce it is a by-product of its outstanding performance on clean CIFAR-100, which shadows its robustness from a relative perspective. Still, it is worth noting that UniCon outperforms other models on most metrics even with a backbone of ResNet-18. Table \ref{tab10} (right) illustrates that UniCon deteriorate less with higher corruption severity. As is analyzed by \cite{poorconv}, though the robustness of deep models can be partly improved by convolutional layers and augmentations, it remains a problem for researchers to handle with better network design, which, in this paper, is partly solved with Mixup-induced universum. Further details can be found in the supplemental material.

\begin{figure}[htbp]
\begin{minipage}[b]{0.45\textwidth}
\includegraphics[width=1\textwidth]{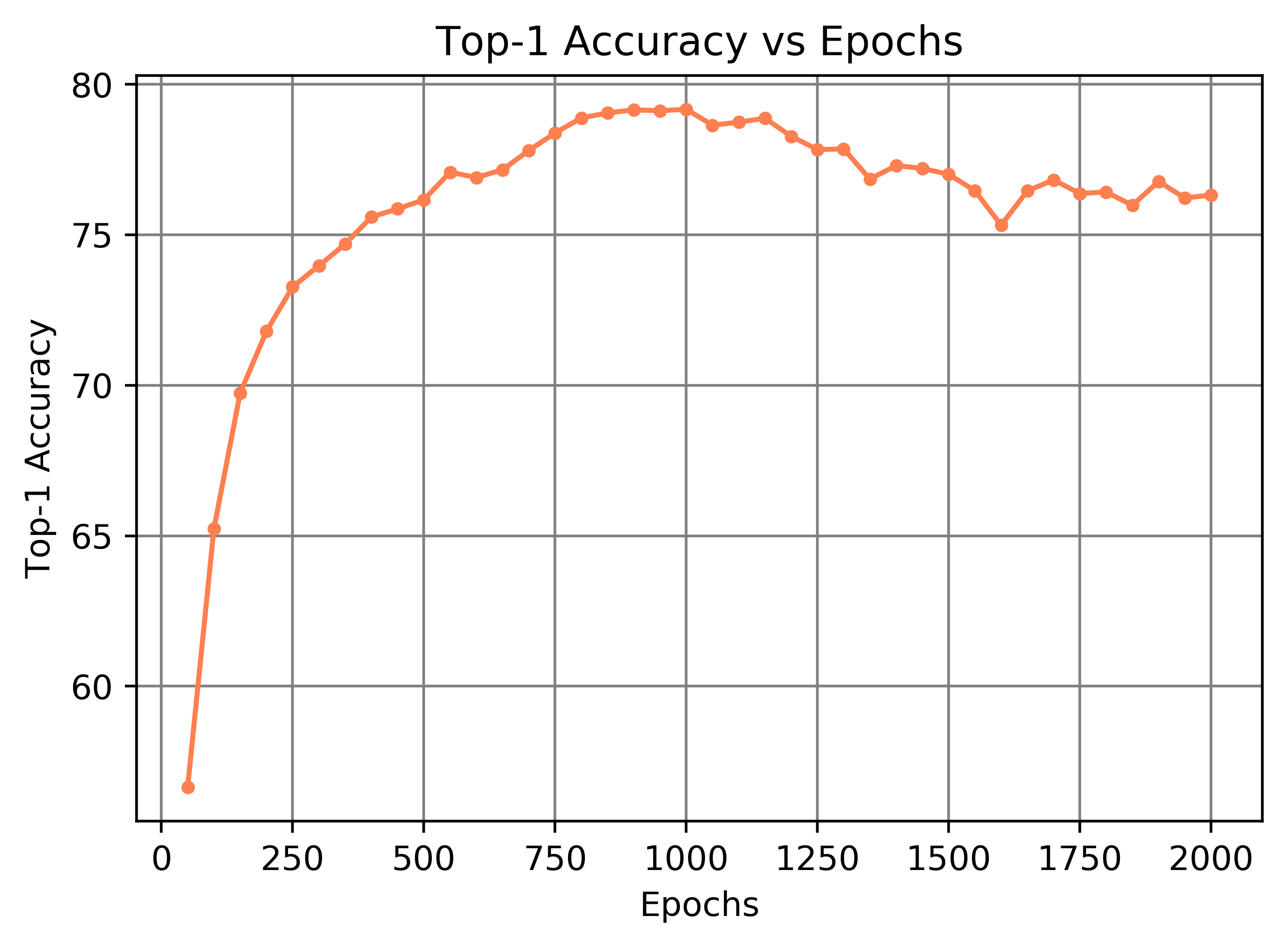}
\caption{Performance of UniCon with different pretraining epochs.} \label{fig5}
\end{minipage}
\end{figure}

\subsection{Hyper-parameter Analysis}\label{parameter}
Fig. \ref{fig3} illustrates UniCon's stability to different hyper-parameters on CIFAR-100. We modify the backbone networks, batch sizes, learning rates and temperature parameter one at a time to observe whether our model is sensitive to the punctuation of hyper-parameters. Generally speaking, UniCon shows promising performance even in the worst situation.

We evaluate our model with a backbone of Resnet-18, Resnet-34, Resnet-50, and Resnet-101, respectively. In the aspect of model sizes, a deeper network would always improve the performance. Specially, UniCon achieves 83.02\% on CIFAR-101. We deduce that stronger networks like PreAct ResNet \cite{preact}, WideResNet \cite{wideresnet} and DenseNet \cite{densenet} can further boost the performance of our model, which is beyond the scope of this paper.

It is worth noting that our model may not always perform better with a larger batch size, as its top-1 accuracy on batch size 512 and 1024 is lower than that on batch size 256. Since a lot of papers have shown that large batch sizes benefit the training of contrastive models \cite{simclr,moco,supcon}, such results can be intriguing. We conjecture that our model, with additional hard negatives generated by Mixup, is a beneficiary of frequent gradient descents. For training epochs of a fixed number, large batches inevitably lead to a decline in optimization times, thereby resulting into worse performance. It is necessary to make a trade-off between large batch sizes and optimization frequencies. As is shown in Fig. \ref{fig3}(b), we find that 256 is the optimal batch size for most cases. 

Fig. \ref{fig5} shows the convergence of UniCon for 2000 epochs. Since cosine annealing we use for learning rate decay is sensitive with different training epochs, for reproductivity we divide the training period into first 1000 epochs and second 1000 epochs, each with a complete process of cosine annealing.

\begin{table}[H]
\setlength{\tabcolsep}{3mm}
\renewcommand\arraystretch{1.2}
\caption{CIFAR-100 classification accuracy for different Mixup Settings. We set $\lambda$ to a constant.}\label{tab2}
\centering
\begin{tabular}{cc}
\hline
$\lambda$ & Top-1 Accuracy\\
\hline
0.3 & 74.7\\
0.4& 76.6\\
0.5& \textbf{79.2}\\
0.6& 77.0\\
0.7& 73.7\\
%$\lambda \sim{\emph{Beta(0.5, 0.5)}}$  & 77.05\\
\hline
\end{tabular}
\end{table}

\begin{table}[H]
\setlength{\tabcolsep}{3mm}
\renewcommand\arraystretch{1.2}
\caption{Sensitivity of $\gamma$ on CIFAR-100.}\label{tab6}
\centering
\begin{tabular}{|c|c|c|c|}
\hline
\multirow{2}{*}{CIFAR-100} & \multicolumn{3}{|c|}{$\gamma$ in Beta sampling} \\
\cline{2-4}
&1.0&0.8&0.5\\
\hline
Acc.(\%)&77.2&77.3&77.0\\
\hline
\end{tabular}
\end{table}

\subsection{Mixup strategies}\label{mixup}
We test different strategies of choosing $\lambda$ in Mixup. We either fix $\lambda$ to 0.3, 0.4, 0.5, 0.6 and 0.7, respectively, or assume that $\lambda$ is a random number subject to $Beta(\gamma, \gamma)$ ($\gamma$ is chosen from 0.5, 0.8 and 1.0), following \cite{mixup}. As Table. \ref{tab2} and Table. \ref{tab6} demonstrates, the model achieves best performance when two images are equally mixed to produce a universum negative. This result is in line with our intuition that the Mixup image is farthest from its original images in semantics when two images make equal contributions to their mixture. In fact, our model benefits from semantically ambiguous images as they make better universum.

Here we present a brief analysis to justify the necessity of setting $\lambda$ to $0.5$. From the perspective of Mixup, the label of a Mixup data point with Mixup parameter $\lambda$ should be $y = \{0,...,\lambda,...1-\lambda,...0\}$, where the positions of $\lambda$ and $1-\lambda$ accord with corresponding mixture data. Therefore, the information entropy of $y$ can be calculated.

\begin{equation}
    H(y) = \lambda log(\lambda) + (1-\lambda)log(1-\lambda)
\end{equation}
To find out the maximal value of $H(y)$, the gradient of $H(y)$ with respect to $\lambda$ is derived.

\begin{equation}
    \frac{\partial H(y)}{\partial \lambda} = log \frac{\lambda}{1-\lambda}
\end{equation}
Apparently, $\frac{\partial H(y)}{\partial \lambda}=0$ when $\lambda=0.5$. At this point, $H(y)$ takes the maximum value, while the label vector $y$ has the highest uncertainty. Consequently, Mixup-induced universum data can be best described as negatives for all when $\lambda=0.5$.

\begin{table}[htbp]
\setlength{\tabcolsep}{1.2mm}
\renewcommand\arraystretch{1.2}
\caption{Ablation study. The loss functions are examined on what data are regarded as negatives for contrast and whether universum data are used for class center derivation.}\label{ablation}
\centering
\begin{tabular}{c|c|c|c|c}
\hline
Loss  & \multicolumn{2}{|c|}{Negatives for contrast} & Class centers &  Top-1\\
\cline{2-3}
function& Universum & Out-of-class&  from universum data& Accuracy\\
\hline
$L_{UniCon}$& \Checkmark & \XSolidBrush &\Checkmark &79.2\\
& \Checkmark& \Checkmark & \Checkmark & 78.3\\
& \XSolidBrush& \Checkmark & \Checkmark & 4.3\\
\hline
$L_{sup}$& \XSolidBrush & \Checkmark & \XSolidBrush & 71.5\\
$L_{add}$& \Checkmark & \Checkmark & \XSolidBrush & 68.7 \\
\hline
\end{tabular}
\end{table}

\subsection{Ablation study}
To further understand the effectiveness of each designed component of our model, an ablation study is conducted. We examine what data are regarded as negatives for contrast and whether universum data are used for class center derivation for each loss. As Table \ref{ablation} demonstrates, it is crucial that universum data are utilized to derive the class centers. The use of additional universum negatives does harm to the model performance, while the mere use of universum for class center derivation will result into deteriorated performance. However, when universum negatives are used in combination with universum-derived class centers, the model acquires the best performance.

\subsection{Performance in the unsupervised setting}
We evaluate the performance of Un-Uni (the unsupervised version of UniCon) on CIFAR-100 with a ResNet-18 backbone. Despite the loss function computation, the implementation details of Un-Uni are the same as UniCon. Un-Uni is compared with SimCLR, Moco-v2\cite{mocov2} and three unsupervised Mixup-boosted contrastive models, that is, Un-Mix\cite{unmix}, MoCHi\cite{hardmix} and Mixco\cite{mixco}.

As is shown in Table. \ref{tab8}, Un-Uni outperforms other counterparts by a small margin. Considering that Un-Uni is only a simple application of UniCon in the unsupervised setting, such performance has proved the potential of our proposed method.  

\begin{table}[htbp]
\setlength{\tabcolsep}{2mm}
\renewcommand\arraystretch{1.2}
\caption{Top-1 classification accuracy (\%) on CIFAR-100 in the unsupervised setting. Our unsupervised model (Un-Uni) is compared with SimCLR\cite{simclr}, Mixco\cite{mixco}, Un-Mix\cite{unmix}, Moco-v2\cite{mocov2} and MoCHi\cite{hardmix}. We use \textbf{bold} to indicate the best result.}\label{tab8}
\centering
\begin{tabular}{ccccc}
\hline
Method &Classifier&  Batch size&Memory size&Acc\\
\hline
SimCLR &Linear&  1024&-& 61.7\\
Mixco &Linear& 256&- & 62.6 \\
Un-Mix&Linear& 256&-& 64.2 \\
\hline
Moco-v2&KNN&512&4096&62.9\\
MoCHi &KNN &512 &4096& 60.8\\
\hline
Un-Uni(ours)&Linear& 256&- & \textbf{64.5}\\
\hline
\end{tabular}
\end{table}

\section{Conclusion}
This paper explores Mixup from the perspective of Universum Learning, thus proposing to assign synthesized samples into a generalized negative class in the framework of supervised contrastive learning. Our model achieves state-of-the-art performance on CIFAR-10, CIFAR-100 and TinyImageNet. The results of our experiments reveal the potential of Mixup to generate hard negative samples, which may open a new window for further studies.

\section*{Acknowledgments}
This research was supported in part by the National Natural Science Foundation of China (62106102, 62076124), in part by the Natural Science Foundation of Jiangsu Province (BK20210292). It is also supported by Postgraduate Research \& Practice Innovation Program of NUAA under No. xcxjh20211601.

\bibliographystyle{IEEEtran}
\bibliography{IEEEabrv,ref}


%\vfill


\end{document}

\onecolumn
\title{Supplemental Material for ``Universum-inspired Supervised Contrastive Learning''}

\author{Aiyang~Han,~Chuanxing~Geng,~Songcan~Chen
}

\markboth{}%
{}

\maketitle

\section*{Details for theoretical analysis}
\subsection{Theoretical analysis}
In Section III-E, we analyze the gradients of UniCon loss to show that it can perform hard negative mining and draw clear margins among different classes in the embedding space. $L_{UniCon,i}$ is defined in the following form.

\begin{equation}
L_{UniCon, i} = -\frac{z_i}{\tau}\cdot m_i +log\sum_{k \neq i}exp(z_i \cdot zu_k/\tau)
\end{equation}

As is shown above, the gradients of $L_{UniCon,i}$ with respect to $z_i$ is calculated.

\begin{equation}\label{gradient}
    \frac{\partial L_{UniCon,i}}{\partial z_i} =
    \frac{1}{\tau}\left[-m_i+\sum_{k \neq i}zu_kPU_k+G\right]
\end{equation}
where we define,
\begin{equation}
PU_k = \frac{exp(z_i \cdot zu_k / \tau)}{\sum_{j\neq i}exp(z_i \cdot zu_j / \tau)}
\end{equation}
\begin{equation}
G = \frac{z_i \sum_{k\neq i} exp(z_i \cdot zu_k / \tau)\frac{\partial zu_k}{\partial z_i}}{ \sum_{k \neq i}exp(z_i \cdot zu_k / \tau)}
\end{equation}

Apparently, as the representation of a class, $m_i$ always has an influence in the optimization process. Moreover, similar to \cite{supcon}, UniCon loss also has the ability of hard sample mining. Considering the second term of Eq.\ref{gradient}, when $zu_k$ is a hard negative, we have $z_i \cdot zu_k \approx 1$. Therefore, 

\begin{align}
PU_k \approx \frac{exp(1 / \tau)}{Z}
\end{align}
which greatly benefits the encoder. In the equations, we define:

\begin{align}
Z = \sum_{j \neq i}exp(z_i \cdot zu_j / \tau)
\end{align}

Otherwise, when $zu_k$ is an easy negative, we have $z_i \cdot zu_k \approx 0$. Therefore, 

\begin{align}
    PU_k \approx \frac{1}{Z}
\end{align}
which narrowly influences the encoder.

Different from $PU_k$ that generally fit into the framework of \cite{supcon}, as a core part of UniCon loss, universum gradient $G$ is unique in forcing large margins among different classes. First we derive a better form of $\frac{\partial zu_i}{\partial z_i}$.

\begin{align}
    \frac{\partial zu_i}{\partial z_i} &= 
    \frac{\partial zu_i}{\partial \widetilde x_i} \cdot \frac{\partial \widetilde x_i}{\partial z_i} \\
    &= \frac{\partial zu_i}{\partial u_i} \cdot \frac{\partial u_i}{\partial \widetilde x_i} \cdot \frac{\partial \widetilde x_i}{\partial z_i} \\
    &= \frac{\partial f(u_i)}{\partial u_i} \cdot \frac{\partial (\lambda \widetilde x_i +(1-\lambda)\widetilde x_{q(i)})}{\partial \widetilde x_i} \cdot \frac{\partial \widetilde x_i}{\partial f(\widetilde x_i)} \\
    &=  \frac{\lambda f^{'}(u_i)}{f^{'}(\widetilde x_i)}
\end{align}
Similarly, we derive a better form of $\frac{\partial zu_k}{\partial z_i}$.

\begin{align}
\begin{split}
\frac{\partial zu_k}{\partial z_i}=\left\{
\begin{array}{cc}
\frac{(1-\lambda)f^{'}(u_k)}{f^{'}(\widetilde x_i)}, &q(k)=i,  \\
0, & otherwise.
\end{array}
\right.
\end{split}
\end{align}
With newly formed $\frac{\partial zu_i}{\partial z_i}$ and $\frac{\partial zu_k}{\partial z_i}$, we rewrite universum gradient $G$ in the following form.

\begin{align}  
  G &= z_i\sum_{k \neq i}PU_k \cdot \frac{\partial zu_k}{\partial z_i} \\
  &=z_i \cdot\sum_{k \in Q_i} \frac{(1-\lambda)f^{'}(u_k)}{f^{'}(\widetilde x_i)}PU_k \\  
    &= \frac{z_i}{f^{'}(\widetilde x_i)}\sum_{k \in Q_i}(1-\lambda)f^{'}(u_k)PU_k
\end{align}
where we denote $Q_i = \left \{k|q(k) = i\right\}$. The subsequent analysis can be found in Section III-E.

\section*{Implementation Details}
\subsection{Implementation of SupCon + Un-Mix}
To further evaluate the performance of SupCon in combination with Mixup, we propose a hybrid model of SupCon and Un-Mix. Following \cite{unmix}, we mix one branch of a batch of training data with the reverse version of itself while maintaining the other branch unchanged. In line with the denotations in Section III, we use $\widetilde x_{2k-1}$ as the Mixup branch and $\widetilde x_{2k}$ as the unchanged branch, where $k=1,2,..,N$. Therefore, the mixed data is acquired through the following equation:

\begin{equation}
    x_{mix,k} = \lambda \cdot \widetilde x_{2k-1} + (1-\lambda) \cdot \widetilde x_{2(N+1-k)-1}, \quad k=1,2,..,N,
\end{equation}
where $\lambda$ is the Mixup parameter sampled from Beta distribution. Unlike \cite{unmix} that uses both Mixup and CutMix with a probability parameter, our implementation only uses Mixup. For clearance, in the following part, we rewrite $L_{sup}$ as $L_{sup}(X, \widetilde X, Y)$, where $X$ and $\widetilde X$ are two branches of data of the same batch size and $Y$ is their corresponding label (two branches share the same label).

\begin{align}\label{eq15}
L_{sup}(X,\widetilde X, Y) =& \sum_{i=1}^{N} \frac{-1}{2|D_i|+1} \{\sum_{d \in D_i} \left[log \frac{exp(z_i\cdot z_d/\tau)}{S}+ log\frac{exp(z_i\cdot \widetilde z_d/\tau)}{S}\right] \\
 &+ log \frac{exp(z_i \cdot \widetilde z_i / \tau)}{S}\}
\end{align}
where $S=\sum_{k \neq i} exp(z_i \cdot z_k / \tau)+\sum_{k} exp(z_i \cdot \widetilde z_k / \tau)$, $D_i \equiv \{k|k\in \{1,2,..,N\}, k \neq i, Y_k= Y_i \}$ a set of indices that refer to samples in the same class with i in each branch, and $z_i=f(X_i)$ and $\widetilde z_i = f(\widetilde X_i)$ are the embeddings of $X$ and $\widetilde X$. With Equation \ref{eq15}, we derive the loss of the combination of SupCon and Un-Mix.

\begin{equation}
    L_{SupMix} = \lambda \cdot \underbrace{L_{sup}(X_{mix}(\downarrow), \widetilde X, Y)}_{normal\ order\ of\ mixtures} + (1-\lambda) \cdot \underbrace{ L_{sup}(X_{mix}(\uparrow), \widetilde X, Y)}_{reverse\ order\ of\ mixtures}
\end{equation}

where $\lambda$ is the Mixup parameter, $X_{mix}$ is the set of $\{x_{mix,k}\}_{k=1,2,...,N}$, $\widetilde X$ is the set of $\{\widetilde x_{2k}\}_{k=1,2,...,N}$, and $Y$ is the label set of $\{\widetilde y_{2k}\}_{k=1,2,...,N}$. Different from \cite{unmix}, $L_{SupMix}$ does not include the original SupCon loss, yet the result is still promising and outperforms SupCon. Although this model is not the main part of this paper, to the best of our knowledge, this is also the first time that traditional Mixup method is combined with supervised contrastive learning.
In the implementation, we empirically adopt the same hyper-parameters as we have for our own model. On CIFAR-100, we test the batch size of 256 and 1024, and observes that just like UniCon, SupMix also benefits from a small batch size of 256 due to the same reason as we have analyzed in Section IV-E.

\subsection{Augment}
For augmentations of our model, we sequentially exert a random cropping, resizing (to input size $32 \times 32$), a random horizontal flip, a random application of color jittering, and a random application of gray scale conversion. The probabilities of application of horizontal flip, color jittering and gray scale conversion are 0.5, 0.8 and 0.2, respectively. In color jittering, the parameters for brightness, contrast, saturation and hue are set to 0.4, 0.4, 0.4 and 0.1, respectively. The augmented images are normalized. When Augment is applied, it is used twice to double the size of the training data no matter whether the model is contrastive. 

\begin{table*}
\UseRawInputEncoding
\begin{lstlisting}[language=Python, title=Algorithm 1: Python-like pseudocode for universum-style Mixup] 
# N: batch size of the training data
# images: an augmented batch of 2N images, the first N images being the first branch and the second N images being the second branch
# labels: the ground truth labels of size N, shared by both branches of the images
# lam: Mixup parameter, set to 0.5 in most cases

def mix_universum(images, labels, lam, N):
    # synthesize the universum negatives from a minibatch
    cls_idx = [[np.where(labels != i)[0]] for i in range(max(labels)+1)]
    chosen_data = [images[random.choice(cls_idx[labels[i % N]])] for i in range(N * 2)]
    mix_data = torch.stack(chosen_data, dim=0)
    universum = lam * images + (1 - lam) * mix_data
    return universum
\end{lstlisting}
\end{table*}

\subsection{A Python-like Code for Universum-style Mixup}
Algorithm 1 presents a Python-like pseudocode of our universum-style Mixup. Unlike the conventional implementation of Mixup and Cutmix that use k-permutations to rearrange the whole batch of data, we choose our Mixup data in a Bootstrapping way. On the one hand, each data point will be used as the anchor for \textbf{once} in each mixture; and on the other hand, the data points can be chosen as a g-negative for more than once or they may not be chosen at all. This strategy allows more randomness. From a statistical perspective, our way can better appropriate the distribution of out-of-class data points, so as to reduce the influence of unbalanced distribution of different classes in a minibatch.

\begin{figure*}[htbp]
    
	\centering
	\subfigure[]{
		\begin{minipage}[b]{0.45\textwidth}
			\includegraphics[width=1\textwidth]{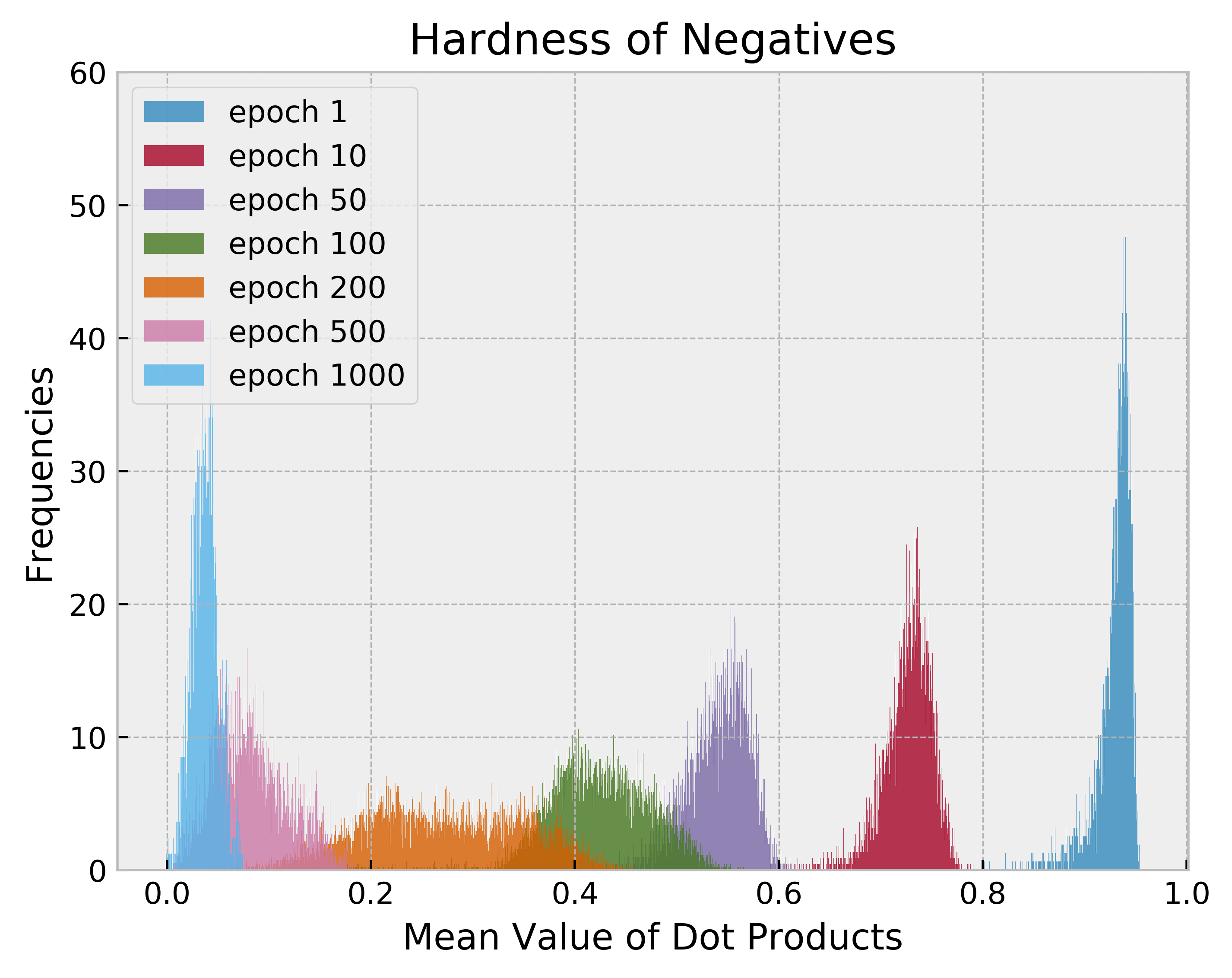} 
		\end{minipage}
	}
    	\subfigure[]{
    		\begin{minipage}[b]{0.45\textwidth}
   		 	\includegraphics[width=1\textwidth]{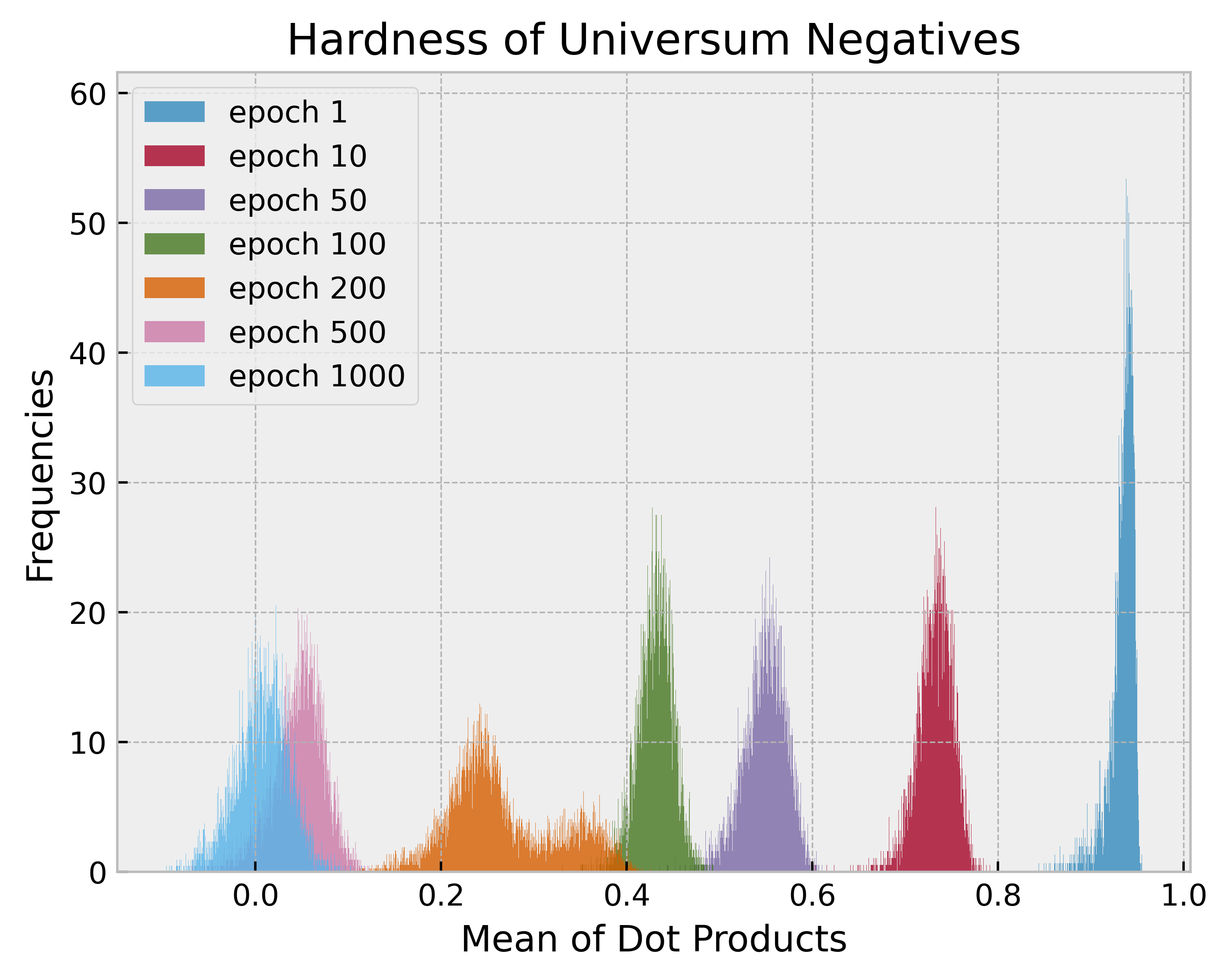}
    		\end{minipage}
    }
	\caption{Histogram of the hardness of out-of-class negatives and universum negatives. }\label{fig6}
\end{figure*}
\section*{Experimental Results}
\subsection{Hardness of Negatives}
The dot product of the anchor feature $z_i$ and its negative feature $z_n$ or universum negative feature $zu_k$ reveals the hardness of negatives. As is analyzed in Section III-D, like most contrastive losses, the loss function of UniCon benefits more from hard negatives. Therefore, it is possible to use hardness to value the contributions of negatives and universum negatives in the process of optimization. Here we denote hardness H as the dot product of an anchor and a negative:

\begin{align}
    H = z_i \cdot z_n \\
    H = z_i \cdot zu_k
\end{align}
with which the hardness of negatives and universum negatives is unified. In the experiment, the features of first 10000 images and their corresponding 10000 universum data points are recorded in epoch [1, 10, 50, 100, 200, 500, 1000] with ResNet-18 on CIFAR-100. We calculate the dot products of each anchor with all of its negatives and universum negatives, drawing a mean value of dot products for each negative data point to denote its hardness. In Fig. \ref{fig6}, the histograms reveal the hardness of traditional out-of-class negatives and universum negatives for different training epochs. Please note that although conventional out-of-class negatives are not directly contrasted in our loss function, the model still automatically recognizes them as negatives. At the beginning of the training, there is little difference between these two kinds of negatives. The hardness of both decreases with training epochs, while the hardness of universum negatives are more concentrated with a higher peak in the middle phase of training. In the middle phase, universum negatives are slightly harder than out-of-class negatives. At the last phase of training, universum negatives become easy, indicating that our assumption of universum data belonging to a generalized negative class which is distinguishable from other classes. 

\begin{table*}[htbp]
\setlength{\tabcolsep}{1.5mm}
\renewcommand\arraystretch{1.5}
\caption{Error on clean datasets, mCE and Corruption Error for different corruptions and models on TinyImageNet-C and CIFAR-100-C. Lower is better for all metrics. The table form and evaluation method is from \cite{corruption}. With this table, we intend to show the robustness of our model in the face of different corruptions.}\label{tab11}
\centering
\begin{tabular}{l|c|c|ccc|cccc|cccc|cccc}
\hline
\multicolumn{18}{c}{\textbf{TinyImageNet-C} } \\
\hline
\multicolumn{3}{c}{} & \multicolumn{3}{c}{Noise} &\multicolumn{4}{c}{Blur} & \multicolumn{4}{c}{Weather} & \multicolumn{4}{c}{Digital}\\
\hline
Network& Err. & \textbf{mCE} &\tiny Gauss. & \tiny Shot &\tiny Impulse&\tiny Defocus&\tiny Glass&\tiny Motion&\tiny Zoom&\tiny Snow&\tiny Frost&\tiny Fog&\tiny Bright&\tiny Contrast&\tiny Elastic&\tiny Pixel&\tiny JPEG \\
\hline
AlexNet&	62.1& 	100.0& 	100&	100&	100&	100&	100&	100&	100&	100&	100&	100&	100&	100&	100&	100&	100\\
Xent-50&	41.7& 	89.9& 	92&	89&	93&	92&	92&	91&	90&	90&	89&	93&	88&	97&	87&	84&	82\\
SupCon-50&	41.7& 	84.9& 	86&	84&	88&	92&	91&	90&	90&	80&	74&	89&	75&	84&	88&	82&	80\\
UniCon-18&	40.7& 	83.2& 	91&	88&	91&	91&	89&	88&	89&	75&	69&	81&	73&	80&	86&	79&	78\\
UniCon-50&	35.0& 	76.4& 	87&	83&	88&	85&	84&	81&	82&	68&	60&	71&	65&	73&	78&	70&	69\\

\hline
\end{tabular}

\begin{tabular}{l|c|c|ccc|cccc|cccc|cccc}
\hline
\multicolumn{18}{c}{\textbf{CIFAR-100-C}} \\
\hline
\multicolumn{3}{c}{} & \multicolumn{3}{c}{Noise} &\multicolumn{4}{c}{Blur} & \multicolumn{4}{c}{Weather} & \multicolumn{4}{c}{Digital}\\
\hline
Network& Err. & \textbf{mCE} &\tiny Gauss. & \tiny Shot &\tiny Impulse&\tiny Defocus&\tiny Glass&\tiny Motion&\tiny Zoom&\tiny Snow&\tiny Frost&\tiny Fog&\tiny Bright&\tiny Contrast&\tiny Elastic&\tiny Pixel&\tiny JPEG \\
\hline
AlexNet&42.9&100&100&100&100&100&100&100&100&100&100&100& 100&100&100&100&100\\
Xent-50	&22.8 &	79.0 &92&88& 79&	88	&72	&91&	98&	67&	73&	65&	57&	71&	75&	90&	79\\
SupCon-50&	24.6& 	84.3& 	106&	101&	91&	65&	141&	88&	69&	84&	80&	66&	54&	50&	80&	99&	90\\
UniCon-18&	20.8& 	73.4& 	106&	103&	99&	56&	117&	67&	61&	59&	62&	51&	46&	36&	69&	81&	87\\
UniCon-50&	18.3& 	70.6& 	107&	105&	96&	51&	124&	62&	56&	54&	57&	43&	41&	31&	62&	84&	86\\

\hline
\end{tabular}
\end{table*}
\subsection{Robustness with Different Corruptions}
In Table \ref{tab11}, error on clean datasets, mCE and Corruption Error for different corruptions and models on TinyImageNet-C and CIFAR-100-C are shown. According to \cite{corruption}, the corruptions can be roughly divided into four categories: noise, blur, weather and digital. On TinyImageNet-C, our models with both ResNet-18 and ResNet-50 backbones significantly outperform counterparts in almost all aspects. On CIFAR-100-C, contrastive models do not perform well in the face of Gaussian noise, shot noise and especially glass blur. Generally, our model can better handle corruptions in the field of blur, weather and digital, while failing to greatly improve the results with noise. Although augmentation techniques inevitably play a role in standing corruptions, our proposed Universum-induced Mixup also helps make the model more robust as UniCon surpasses SupCon with the same augmentations by a great margin on mCE.

\bibliographystyle{IEEEtran}
\bibliography{IEEEabrv,ref}